\documentclass[pdflatex,sn-nature]{sn-jnl}

\usepackage{graphicx}%
\usepackage{multirow}%
\usepackage{amsmath,amssymb,amsfonts}%
\usepackage{amsthm}%
\usepackage{mathrsfs}%
\usepackage[title]{appendix}%
\usepackage{xcolor}%
\usepackage{textcomp}%
\usepackage{manyfoot}%
\usepackage{booktabs}%
\usepackage{algorithm}%
\usepackage{algorithmicx}%
\usepackage{algpseudocode}%
\usepackage{listings}%
\usepackage{makecell}

\usepackage{siunitx}
\theoremstyle{thmstyleone}%
\theoremstyle{thmstyletwo}%

\theoremstyle{thmstylethree}%

\begin{document}

\title[Article Title]{BenchECG and xECG: a benchmark and baseline for ECG foundation models}


\author[1]{\fnm{Riccardo} \sur{Lunelli}}
\equalcont{These authors contributed equally to this work.}

\author[1]{\fnm{Angus} \sur{Nicolson}}
\equalcont{These authors contributed equally to this work.}

\author[1]{\fnm{Samuel Martin} \sur{Pröll}}

\author[2]{\fnm{Sebastian Johannes} \sur{Reinstadler}}
\author[2]{\fnm{Axel} \sur{Bauer}}

\author*[1]{\fnm{Clemens} \sur{Dlaska}}\email{clemens.dlaska@i-med.ac.at}

\affil[1]{\orgdiv{Digital Cardiology Lab, University Clinic of Internal Medicine III}, \orgname{Medical University Innsbruck}, \orgaddress{\state{A-6020 Innsbruck}, \country{Austria}}}
\affil[2]{\orgdiv{University Clinic of Internal Medicine III, Cardiology and Angiology}, \orgname{Medical University Innsbruck}, \orgaddress{\state{A-6020 Innsbruck}, \country{Austria}}}



\abstract{
Electrocardiograms (ECGs) are inexpensive, widely used, and well-suited to deep learning. Recently, interest has grown in developing foundation models for ECGs -- models that generalise across diverse downstream tasks. 
However, consistent evaluation has been lacking: prior work often uses narrow task selections and inconsistent datasets, hindering fair comparison. 
Here, we introduce BenchECG, a standardised benchmark comprising a comprehensive suite of publicly available ECG datasets and versatile tasks. We also propose xECG, an xLSTM-based recurrent model trained with SimDINOv2 self-supervised learning, which achieves the best BenchECG score compared to publicly available state-of-the-art models. In particular, xECG is the only publicly available model to perform strongly on all datasets and tasks.
By standardising evaluation, BenchECG enables rigorous comparison and aims to accelerate progress in ECG representation learning. xECG achieves superior performance over earlier approaches, defining a new baseline for future ECG foundation models.
}

\keywords{ECG, Foundation Model, Benchmark, Self-supervised learning}



\maketitle

Cardiovascular diseases (CVDs) are the leading global cause of death, and early and accurate diagnosis is critical to reducing their burden \cite{Lindstrom2022GlobalBurdenCardiovascular}. Electrocardiograms (ECGs) are inexpensive, non-invasive, and widely available biosignals that integrate complex biological information of the entire organism and are therefore used to detect a broad spectrum of cardiac and non-cardiac conditions. These characteristics make ECGs not only central to routine care but also attractive candidates for automated analysis using machine learning.

\begin{figure*}[t]
    \includegraphics[width=\textwidth]{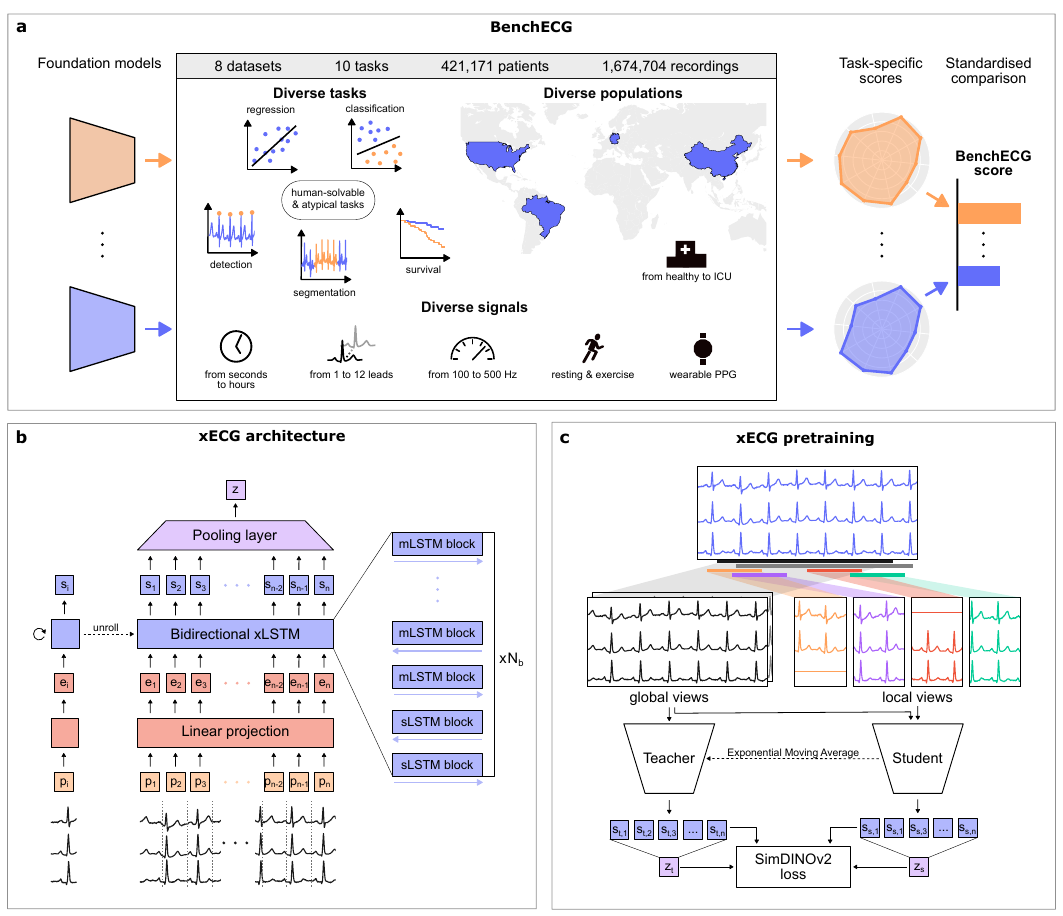}
    \centering
    \caption{\textbf{Overview of BenchECG and xECG.} BenchECG (\textbf{a}) provides a comprehensive evaluation of ECG foundation models with diverse signals, populations and tasks. xECG (\textbf{b}) is a bi-directional recurrent model based on the xLSTM~\citep{beck2024xlstm} architecture and pretrained via SimDINOv2~\citep{wu2025simplifying} self-supervised pretraining (\textbf{c}).}
    \label{fig:fig1}
\end{figure*}

While deep learning models have demonstrated strong performance in interpreting ECGs, when trained on large, labelled datasets \cite{ribeiro2020automatic, Li2025Founder}, the emergence of foundation models represents a paradigm shift for medical artificial intelligence (AI) \citep{Moor2023FoundationModels}. Foundation models are trained on vast, diverse datasets -- often unlabelled -- and designed to be general-purpose and adaptable to multiple downstream tasks.  Recently, several works have proposed foundation models for ECGs that aim to support a variety of downstream tasks across datasets, tasks, and signals \citep{han2024foundationmodelselectrocardiogramreview, Na2024ST-MEM, Kim20204ECGJEPA, Li2025Founder}. These models typically use self-supervised learning (SSL) approaches and transformer-based architectures \cite{Na2024ST-MEM, Kim20204ECGJEPA} that can be trained on vast unlabelled datasets. 

Despite promising results, the field faces two major limitations. First, existing evaluation practices are inconsistent, reflecting a broader ``reproducibility crisis'' in machine learning applied to the sciences \cite{Henderson2018DeepReinforcement, Pineau2020ImprovingReproducibility, Roberts2021CommonPitfalls, Kapoor2023LeakageReproducibilityCrisis, Ball2023AILeadingReproducibility}. Prior work varies significantly in which tasks and datasets are used for validation, making it difficult to compare methods or identify generalisable approaches \cite{Bernardini2021AIOSASleepApnea, han2024foundationmodelselectrocardiogramreview}. Data leakage, improper data preprocessing, and high computational costs often hinder independent replication and verification of results \cite{Kapoor2023LeakageReproducibilityCrisis}. Given the high-stakes nature of medical diagnostics, there is a critical need for trustworthy and transparent AI evaluation. Second, as we show in this work, the architectural choice of using transformer-based models limits performance in long-context tasks, relevant for real-world settings involving long-term monitoring. While, in principle, powerful for capturing long-range dependencies, standard transformers scale quadratically in time and memory complexity with respect to sequence length, making them computationally expensive and memory-intensive for very long time-series data -- a significant limitation given that important clinical applications require analysis of up to hours or days of ECG signals.

A wide range of machine learning fields have addressed similar challenges through the development of standardised, public benchmarks. From well-known benchmarks in vision \cite{Deng2009ImageNet, Lin2014MicrosoftCOCO} or language \cite{wangMMLUProMoreRobust2024, bosmaDRAGONBenchmarkClinical2025}, to more specific applications in molecular sequences \cite{Ektefaie2024EvaluatingGeneralizability}, and federated learning \cite{Karargyris2023FederatedBenchmarkingMedical}. These benchmarks have enabled robust evaluation, driven architectural innovation, and allowed for rapid progress. In this work, we argue that ECG foundation models would benefit from a similar ecosystem. To this end, we introduce \textbf{BenchECG}, a comprehensive benchmark that spans eight publicly available datasets and ten diverse and representative tasks. BenchECG provides a unified platform for evaluating generalisation, task diversity, and representation quality in ECG foundation models.

We also introduce \textbf{xECG}, a novel ECG foundation model based on the extended long short-term memory (xLSTM) architecture~\cite{beck2024xlstm}. xLSTM models combine the benefits of transformers (parallelisable networks that improve with scale) with the efficiency of recurrent structures, allowing for processing of longer ECG sequences, and have recently been shown to excel at signal forecasting~\cite{Auer2025TiRex, plstm2025}. We pretrain xECG using SimDINOv2~\citep{wu2025simplifying}, a state-of-the-art SSL method originally developed for computer vision, adapted here to the time-series domain. We show that this combination yields strong, general-purpose ECG representations, providing out-of-the-box flexibility covering all recording scenarios present in cardiac monitoring.

Across BenchECG, xECG achieves the best average rank when finetuning ($1.50$) and under linear probing ($1.20$), outperforming prior publicly available state-of-the-art models. xECG is the only model to perform strongly across all BenchECG datasets and task types.

In summary, the contributions of this paper are:

\begin{itemize}
    \item We introduce BenchECG, the first comprehensive, standardised benchmark for ECG foundation models, enabling rigorous and reproducible evaluation
    \item We propose xECG, an ECG foundation model that combines the efficiency of xLSTMs with the representational strength of SimDINOv2 self-supervised learning
    \item We demonstrate that xECG achieves state-of-the-art performance across diverse ECG tasks, establishing a strong and reproducible baseline for future work
    \item We release code, models, and benchmark tasks to support open research in general-purpose ECG representation learning
\end{itemize}

\subsection*{A standardised benchmark with clinically relevant tasks}
Foundation models, by definition, should be adaptable to many downstream tasks \cite{Bommasani2021OpportunitiesFoundationModels}. Hence, we require ECG foundation models to handle variety in three different forms: (1) signal characteristics, (2) dataset characteristics, and (3) task characteristics. An ECG foundation model should be evaluated not only on conventional 12-lead clinical ECGs but also on long-term, few-lead, and related (but non-ECG) physiological signals, across geographically and clinically distinct cohorts, and over a broad set of clinically relevant downstream tasks. To this end, we introduce \textbf{BenchECG}, an open benchmark designed to test ECG foundation models made of eight publicly available datasets, comprising $421,171$ patients, and a total of $1,674,704$ recordings (for an overview see Fig.~\hyperref[fig:fig1]{1a} and Table~\ref{tab:BenchECG_datasets}).

BenchECG encompasses signal types used in cardiovascular diagnosis, monitoring and beyond (see Table~\ref{tab:BenchECG_datasets}). Besides short 12-lead ECGs (PTB-XL~\citep{wagner2020ptb}, CPSC2018~\citep{liu2018CPSC}, MIMIC-IV-ECG~\citep{gow2023mimic, goldberger2000physiobank, johnson2023aMIMICIV, johnson2023bMIMICIV}, CODE-15$\%$~\citep{ribeiro_2021_4916206}), longer recordings are represented with 30-minute 2-lead signals (MIT-BIH~\citep{moody2001impact}) and single-lead overnight recordings (Apnea-ECG~\cite{chang2020sleep}). By sourcing data from many datasets we also encounter different sampling rates ($100$-$500$\,Hz) as well as different recording conditions ranging from routine ambulatory settings to high-intensity exercise examinations (Exercise-ECG~\citep{degiovanni2021exercise}). In addition, wearable photoplethysmography (PPG) waveforms (DeepBeat~\citep{torres-sotoMultitaskDeepLearning2020}) are included to test generalisation to other cardiac biosignals.

BenchECG includes cohorts from distinct geographies (Europe, USA, China, Brazil) and populations (healthy individuals to critically ill patients). The datasets also vary in size with the number of participants spanning five orders of magnitude ($20$ to $233,770$ individuals). This ensures the evaluation of models in both low-data and high-data scenarios.

Foundation models should be able to complete a wide variety of downstream tasks, yet key previous works solely evaluate on classification tasks \cite{Kim20204ECGJEPA, Na2024ST-MEM}. To provide a more comprehensive picture of model performance, BenchECG includes different types of clinically relevant tasks:

\begin{itemize}
    \item \textbf{Classification}: multilabel classification of diagnostic labels (PTB-XL, CPSC2018), heartbeat-level arrhythmia classification (MIT-BIH), AF classification from PPG signals (DeepBeat), and multilabel classification for simultaneous (ab)normality assessment of various blood test values (MIMIC-IV-ECG).
    \item \textbf{Segmentation}: sleep apnea segmentation in overnight ECG recordings (Apnea-ECG).
    \item \textbf{Detection}: R-peak detection in standard ECGs (MIT-BIH) and exercise ECGs (Exercise-ECG).
    \item \textbf{Regression}: age estimation across populations (finetuned on CODE-15\%, evaluated on PTB-XL, CPSC2018 and MIMIC-IV-ECG)
    \item \textbf{Survival analysis}: mortality risk prediction across populations (finetuned on CODE-15\%, evaluated on MIMIC-IV-ECG)
\end{itemize}

By encompassing both \textit{typical} tasks (i.e., human-solvable tasks like arrhythmia classification, R-peak detection) and \textit{atypical} (i.e., machine-learning-enabled tasks and tasks traditionally not ECG-associated like laboratory value classification and age estimation), BenchECG provides a rigorous test bed for general-purpose ECG representation learning. 
The evaluation of generalisability is further strengthened by specifically testing out-of-distribution (OOD) performance. For instance, the age estimation task is finetuned on a Brazilian population and evaluated on datasets from different countries (China, Germany and the USA).
Population and prevalence shifts are also accounted for in the mortality risk prediction task where models are finetuned on a general population cohort (CODE-15\%, $5$-year mortality $4.7\%$) and evaluated on ICU patients (MIMIC-IV, $5$-year mortality $36\%$).
Furthermore, models are pretrained on resting ECGs, whereas evaluation on the high intensity exercise ECG task assesses generalisation across substantially different signals.
Finally, in the AF classification from PPG task, we can test a model's adaptability to a completely new modality.

For a fair comparison, models that wish to be evaluated on BenchECG should not be pretrained on any of the evaluation datasets (Table~\ref{tab:BenchECG_datasets}). Note that the CODE-15\% dataset is not used in BenchECG evaluation, but is used during finetuning for the age estimation and mortality prediction tasks. This is because it is commonly used in pretraining of self-supervised methods~\citep{Na2024ST-MEM, Kim20204ECGJEPA} and is a subset of the CODE dataset~\citep{ribeiro2019tele}, used in xECG pretraining.

\subsection*{xECG}

We introduce xECG, a novel ECG foundation model based on the recently introduced extended long short-term memory (xLSTM) architecture~\citep{beck2024xlstm}. Designed to overcome key limitations of transformer-based models, xECG enables efficient processing of long ECG sequences, while retaining strong representational capacity through self-supervised pretraining. The model architecture and training procedure are shown in Figure~\hyperref[fig:fig1]{1b} and \hyperref[fig:fig1]{1c}.

Standard long short-term memory (LSTM) models have limitations in capacity and scalability: limited storage due to scalar cell states, sequential dependencies that prevent parallelism, and rigid gating dynamics that hinder memory flexibility \cite{beck2024xlstm}. Transformers \citep{NIPS2017_3f5ee243} address these constraints but introduce quadratic complexity in sequence length, which is impractical for high-resolution ECG signals recorded over long durations. To address this, xLSTM uses both the scalar (sLSTM) and matrix (mLSTM) memory blocks, as well as exponential gating and multi-head memory mechanisms introduced by Beck et al. \citep{beck2024xlstm}. These innovations allow xLSTMs to scale linearly in sequence length and, unlike standard LSTMs, support training-time parallelism.

The xLSTM architecture has demonstrated strong performance in various sequence modelling tasks \cite{Auer2025TiRex, plstm2025}, including bio-signal generation \cite{schmidinger2025bioxlstm} and ECG analysis \cite{kang2025xlstm}. This extends to computer vision, where the Vision-LSTM (ViL) \cite{alkin2025visionlstm} architecture processes image patches bidirectionally with mLSTM blocks, achieving a better performance-to-cost trade-off than transformer-based approaches. 

We designed xECG with a stack of alternating sLSTM and mLSTM layers arranged bidirectionally. This contrasts with mLSTM-only designs such as ViL \cite{alkin2025visionlstm}, as recent evidence suggests a mixed block architecture improves ECG modelling \cite{kang2025xlstm}.
Specifically, at each layer, one block processes the sequence forward and the other in reverse -- see Figure~\hyperref[fig:fig1]{1b}. This design allows the architecture to aggregate information efficiently both forwards and backwards in time.

Each input ECG is divided into non-overlapping temporal patches and projected into an embedding space before being passed to the xLSTM encoder. These patch-level representations can then be flexibly adapted for downstream tasks: pooled for signal-level classification and regression tasks, or directly used for beat-level classification, detection and segmentation tasks.

Supervised models are limited by the availability and quality of labelled ECG data. In contrast, self-supervised learning (SSL) enables models to scale to much larger unlabelled datasets by generating supervisory signals directly from the data \citep{Bommasani2021OpportunitiesFoundationModels}. Different SSL methods use different strategies: SimCLR \citep{Chen2020SimCLR} contrasts augmented versions of the same signal against others in the batch; Masked Data Modeling (MDM) methods, like Masked Autoencoders \citep{He_2022_CVPR, Na2024ST-MEM}, reconstruct masked portions of the input; and Joint Embedding Predictive Architectures (JEPA) \citep{assran2023self, Kim20204ECGJEPA} mask features and predict representations in latent space. Features learnt by SSL tend to be more robust across tasks, particularly in the low-data regime, when compared to supervised methods as they do not overfit on the specific labels used in training \citep{ericssonHowWellSelfSupervised2021}. These properties make SSL an attractive pretraining strategy for ECG foundation models designed to generalise across diverse signals, patient populations and tasks.

To pretrain xECG, we adopt an SSL approach using SimDINOv2 \citep{wu2025simplifying}, a recent variant of the self-distillation with no labels (DINO) framework \cite{Caron_2021_ICCV, oquab2024dinov2} designed for training stability and reduced hyperparameter sensitivity. Previous work in the medical domain using SSL has typically relied on contrastive methods~\cite{Chen2020SimCLR, Huang2023SelfsupervisedLearningMedical, Liu2023SelfSupervisedContrastiveLearning}, which require large batch sizes and careful augmentation design. By contrast, SimDINOv2 is based on a non-contrastive teacher-student architecture, in which the student network learns to match representations produced by a slowly evolving teacher model.

In this work, we adapt SimDINOv2 to ECG signals and apply a similar multi-view strategy: each training sample is augmented into multiple global and local views, reflecting different temporal crops and physiologically realistic perturbations. However, by including pretraining datasets that have multiple ECGs per patient, we can use global and local views across different signals from the same patient. The student is trained to produce consistent embeddings across these views, while maintaining high feature diversity via a coding rate regulariser. 

The result is a scalable, general-purpose ECG encoder capable of producing rich representations across a wide range of time scales, patient populations, and downstream tasks. When evaluated on BenchECG, xECG achieves the highest overall performance, setting a new baseline for ECG foundation models.

\begin{table*}[t]
\label{tab:BenchECG_datasets}
{\footnotesize
\begin{tabular}{p{2.5cm}ccccp{4.5cm}}
\toprule
\textbf{Dataset}& Leads & Participants	& Recordings  &  Length & Task(s) \\
\midrule
Apnea-ECG~\cite{chang2020sleep}& 1& 70& 70&  7-10 h& Segmentation of signal into apnea vs. non-apnea\\
CPSC2018~\citep{liu2018CPSC}& 12& 6,877&  6,877&6-60 s& Multilabel classification of diagnostic classes; age regression (OOD)  \\
DeepBeat \citep{torres-sotoMultitaskDeepLearning2020}& 1&   169& 500,000 &25 s& Photoplethysmography AF classification (OOD) \\
Exercise-ECG \citep{degiovanni2021exercise}& 1&   20& 100&20 s& R-peak detection under exercise conditions (OOD)\\
MIMIC-IV-ECG \citep{gow2023mimic, goldberger2000physiobank} & 12 & 161,352 & 800,035 & 10 s & Multilabel classification of diagnostic classes, age regression (OOD); Blood test multilabel (a)bnormality classification; survival analysis (OOD) \\
MIT-BIH~\citep{moody2001impact}& 2& 44& 44&30 min& Multilabel heartbeat-level arrhythmia classification; R-peak detection \\
PTB-XL~\citep{wagner2020ptb}& 12& 18,869& 21,799&10 s& Multilabel classification of diagnostic classes; age estimation \\
\midrule
CODE-15\% & 12 & 233,770 & 345,779 & 7-10 s& Training set for age regression and survival analysis. \\
\botrule
\end{tabular}}
\caption{\textbf{BenchECG datasets and tasks.} BenchECG includes a diverse suite of publicly available datasets varying in signal length, number of leads, and task type. Tasks explicitly addressing out-of-distribution aspects are highlighted as OOD. The number of PPGs provided in the DeepBeat dataset include augmentations. CODE-15\% is exclusively used for training as it is included in pretraining datasets.}\label{tab:BenchECG_datasets}
\end{table*}

\section*{Results}\label{sec: Results}

In addition to xECG, we evaluate state-of-the-art ECG foundation models with publicly available pretrained weights on BenchECG: ST-MEM~\citep{Na2024ST-MEM}, ECG-JEPA~\citep{Kim20204ECGJEPA}, and ECGFounder~\citep{Li2025Founder}. Further ablations are conducted using a transformer pretrained via our xECG pretraining strategy (SimDINOv2 Transformer) and a supervised xLSTM. (see \hyperref[sec: Model Selection]{\color{black} Methods} for an overview of each model).

For each method, we train the models five times (via both finetuning and linear probing for pretrained models) with different random seeds for batch collection and linear head initialisation, and we report the mean and standard deviation across runs. For CODE-15\% and Sleep-Apnea-ECG, which have no published validation split, we used a different random train/validation split for each run, but for the other datasets we used the validation split consistent with the literature (see Methods for full details).

\begin{figure*}[t]
    \includegraphics[width=\textwidth]{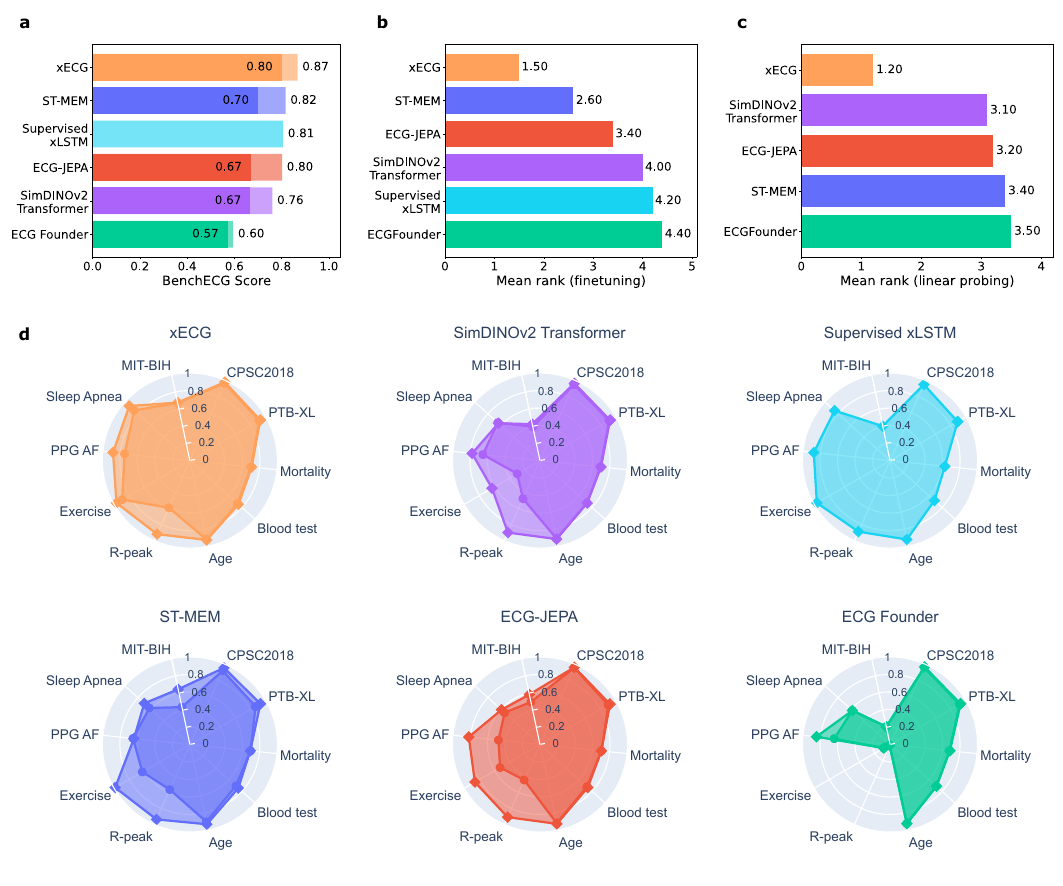}
    \centering
    \caption{\textbf{Performance of state-of-the-art ECG foundation models.} \textbf{a}, BenchECG score for state-of-the-art ECG foundation models (and supervised xLSTM baseline) evaluated using finetuning (light colour) and linear probing (full colour). \textbf{b,c,} Mean rank of evaluated models across BenchECG tasks for finetuning (\textbf{b}) and linear probing (\textbf{c}). \textbf{d}, Radar plots showing performance of models on individual tasks, both for fine-tuning (diamonds) and linear probing (circles).}
    \label{fig:fig2}
\end{figure*}

\subsection*{Comparing foundation models - BenchECG Score}

To compare different foundation models, we propose the \textbf{BenchECG score}: the mean performance of a model across all tasks in the BenchECG benchmark. For each task we select a metric normalised to the range of $0$ to $1$. We use area under the receiver operator characteristic curve (AUROC) for classification\footnote{Apart from for the MIT-BIH task where we follow previous works and use F1 score \cite{xia2023transformer,sellami2019robust, li2022inter, marinho2019novel, li2019automated}.} and segmentation tasks, the symmetric mean absolute percentage error (SMAPE) for regression, F1 score for detection tasks, and the concordance index (C-index) for the survival analysis task (for a detailed justification of metric choices see Methods).

Figure~\hyperref[fig:fig2]{2a} shows the BenchECG score for the different foundation models. xECG achieves the highest BenchECG score of $0.868 \pm 0.0030$, corresponding to an average rank of $1.50$. Results for individual tasks are available in Figure~\hyperref[fig:fig2]{2d} and detailed in Supplementary information. For ranking purposes, pairwise differences between models on the per-task evaluation metrics were assessed using two-sided Welch's t-tests across independent runs, with significance defined as $p<0.05$.

\subsection*{xLSTMs outperform in long-context tasks}

Given the recurrent nature of xLSTMs and their ability to efficiently process longer signals, they are suited for tasks requiring extended temporal context than transformer- or CNN-based models, which here were limited to fixed $10s$ inputs. This is apparent in the sleep apnea task, where understanding the variability in a patient's ECG over long time periods is crucial. The supervised xLSTM achieved a higher AUROC than the next best foundation model, ST-MEM ($0.853\pm0.022$ vs. $0.702\pm0.020$, $p=0.000004$), while xECG achieved the highest AUROC overall ($0.932\pm0.014$), significantly outperforming ST-MEM ($p=0.0000001$) and the supervised xLSTM ($p=0.0003$).

The MIT-BIH arrhythmia classification task also involves long-context inputs with $30$-minute ambulatory ECGs. In terms of F1 score, ST-MEM outperforms prior methods reported in the literature \citep{xia2023transformer, sellami2019robust, li2022inter, marinho2019novel, li2019automated}, but is itself surpassed by xECG ($0.644\pm0.007$ vs. $0.677\pm0.025$, $p=0.040$, comparing ST-MEM to xECG). This highlights the benefit of large-scale self-supervised pretraining. However, the gap is larger under linear probing, where xECG achieves an F1 score of $0.674\pm0.013$ compared to $0.436\pm0.036$ for ST-MEM ($p=0.000032$), indicating that the pretrained xLSTM features capture more relevant information for long-context tasks. Supplementary Table~\ref{tab:mitbih_results} reports the scores for all methods.

\subsection*{Foundation models solve standard evaluation tasks}

On shorter 12-lead clinical recordings, which are the focus of most prior work \cite{Na2024ST-MEM, Kim20204ECGJEPA, Li2025Founder}, performance differences between models were narrower. On PTB-XL, all foundation models were within $0.008$ AUROC of each other (between $0.923$ and $0.931$) after finetuning, while on CPSC2018 the spread was wider at $0.023$ (between $0.958$ and $0.981$), with xECG achieving the highest score (AUROC $0.981$). There were substantially larger gaps between models when linear probing, with AUROCs varying between $0.867$ and $0.917$ for PTB-XL and $0.922$ and $0.968$ for CPSC2018 -- a difference of $0.050$ and $0.046$, respectively.
For the complete set of scores see Supplementary Tables~\ref{tab:ptb_xl_results} and~\ref{tab:cpsc2018_results}. 

\subsection*{Generalisation across populations}
\begin{figure*}[t]
    \includegraphics[width=\textwidth]{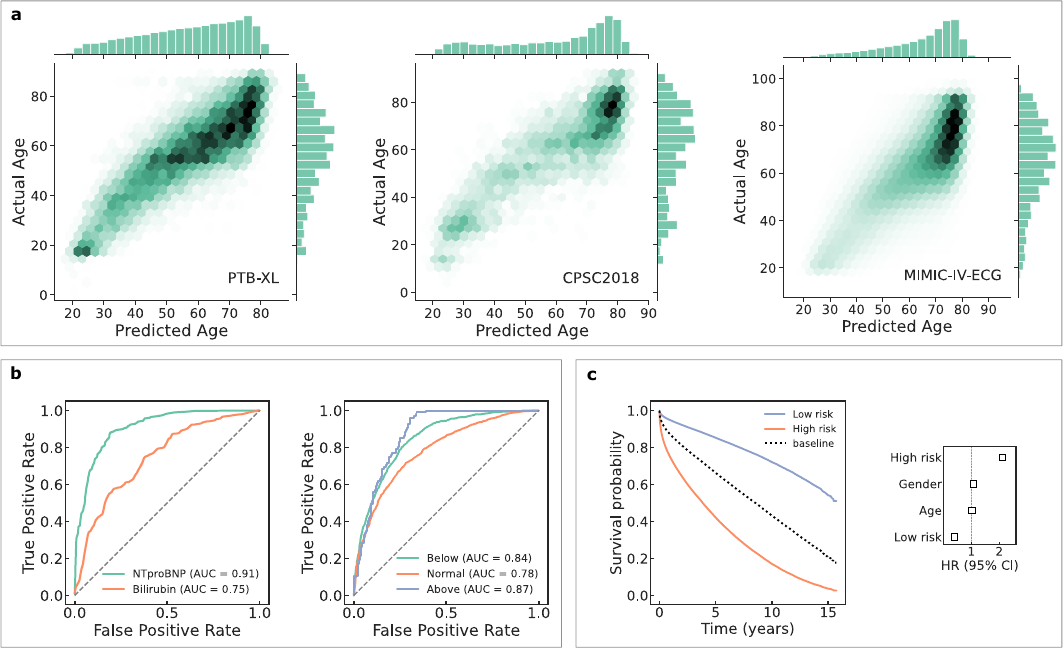}
    \centering
    \caption{\textbf{Additional results for xECG}. \textbf{a,} The distribution of age predictions on the three test sets. \textbf{b,} Performance for NT-proBNP and Bilirubin normal/abnormal classification (left) alongside Albumin abnormally low/normal/abnormally high classification (right) on MIMIC-IV-ECG. \textbf{c,} Kaplan Meier curves and hazard ratios for the age/sex adjusted mortality risk on MIMIC-IV-ECG.}
    \label{fig:fig3}
\end{figure*}

To test cross-population generalisation, BenchECG includes tasks where models were finetuned and evaluated on different datasets. For the age estimation task, each model was trained on CODE-15\% and tested on MIMIC-IV-ECG, PTB-XL and CPSC2018. These datasets are collected from Brazil, the USA, Germany, and China, respectively. ST-MEM has the lowest mean absolute error (MAE) across all test sets (mean MAE $8.546 \pm 0.070$ years), although xECG has the smallest generalisation gap (difference between validation performance on CODE-15\% and mean test performance) at $1.77$ years -- see Figure~\hyperref[fig:fig3]{3a} for prediction distributions.

For the survival analysis task, models were finetuned on CODE-15\%, a general patient population in Brazil ($5$-year mortality of $4.7\%$) and evaluated on MIMIC-IV-ECG, an ICU population in the USA ($5$-year mortality of $36\%$). This tests each model's ability to generalise across different patient populations beyond country of origin. In all cases, the C-index decreased for MIMIC-IV-ECG compared to CODE-15\%, but remained above $0.6$, suggesting the models retain some predictive value across populations.

To demonstrate the mortality task's clinical relevance, we performed an additional analysis on risk predictions of the xECG model -- see Figure~\hyperref[fig:fig3]{3c}. Patients in the MIMIC-IV-ECG dataset were split into high risk, baseline, and low risk groups using the $25$th and $75$th percentile of the model's risk scores. Adjusting for the age and sex, we fit a cox proportional hazards model and found a hazard ratio of $0.389$ ($0.383$-$0.395$ $95\%$CI) and $2.10$ ($2.09$-$2.12$ $95\%$CI) for the low and high risk groups, respectively. Meaning the high risk group is at twice as much risk as the baseline.

\subsection*{Generalisation across tasks}
Most of the foundation models adapted well to PPG, an ECG-related modality they were not pretrained on.
However, there were large differences in behaviour across models. For example, ST-MEM performed well under linear probing, ranking $2$nd, but saw no improvement upon finetuning and had the lowest AUROC, significantly lower than the second-worst model, SimDINOv2 Transformer (AUROC $0.643 \pm 0.049$ vs $0.780 \pm 0.014$, $p=0.0025$). In contrast, ECG-JEPA's frozen features were equivalent to random predictions at AF detection when linear probing (AUROC of $0.477 \pm 0.006$), but substantially improved after finetuning (AUROC of $0.818 \pm 0.021$). 
Supplementary Table~\ref{tab:ppg_results} summarises the scores for all methods.

For the blood test prediction task performance was limited, with a highest mean AUROC of $0.747$ (for xECG). The only blood test result exceeding an AUROC of $0.8$ was the prediction of normal/abnormal NT-proBNP, a biomarker used in diagnosing heart failure. This leaves room for improvement for future foundation models. Supplementary Table~\ref{tab:lab_test_results} summarises the scores for all methods.

R-peak detection was finetuned and evaluated on two different datasets: MIT-BIH and Exercise-ECG. If we report F1 score for whether the R-peak predictions are within $150$ms of the ground truth, as in previous work \cite{aliReliableECGAnalysis2024}, the transformer and xLSTM-based models are near-perfect with F1 scores greater than $0.99$. Hence, we propose using a more challenging metric, requiring the models to be more precise with their estimates, by reducing the window for a correct R-peak to $20$ms. ECGFounder, as a CNN-based foundation model performed poorly (F1 score of $0.011 \pm 0.001$ and $0.083 \pm 0.002$ for MIT-BIH and Exercise-ECG, respectively). In contrast, ST-MEM and xECG performed exceptionally well with an F1 greater than $0.9$ on both datasets, especially considering the training set for Exercise-ECG consists of only eight patients.

\subsection*{xECG features require less tuning}

Linear probing experiments consistently demonstrated that xECG features generalise better across tasks compared to the other foundation models, with xECG ranking first in all tasks ($p<0.05$), except mortality risk prediction and PTB-XL, where it ranked second. This led to a mean rank of $1.2$ across the benchmark. This advantage is particularly clear when comparing the difference in BenchECG score for finetuning and linear probing for each method (Figure~\hyperref[fig:fig2]{2a}). 

\subsection*{Computational Cost}

\begin{figure*}[t]
    \centering
    \includegraphics[width=\textwidth]{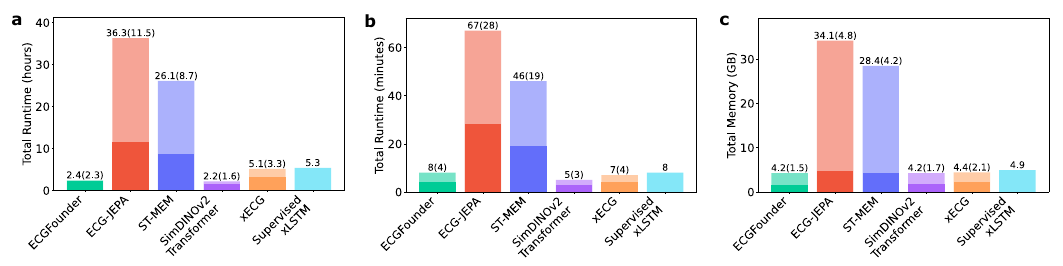}
    \caption{\textbf{Runtime and memory consumption for task-specific adaptation of foundation models}. Total runtime required for finetuning (shaded color) and linear probing (full color) across all BenchECG tasks (\textbf{a}) and PTB-XL task (\textbf{b}) on a single Nvidia L40S GPU. (\textbf{c}) Corresponding memory consumption of different models for the PTB-XL task.}  
    \label{fig:fig4}
\end{figure*}

Finally, we compared computational efficiency. Transformer-based models scale quadratically with input length, making them costly to apply to long ECGs. In contrast, xLSTMs scale linearly, allowing them to process longer recordings with substantially reduced memory and time requirements. On the same hardware\footnote{A single L40S Nvidia GPU}, xECG required $\sim5\times$ less finetuning time across the benchmark than ST-MEM ($5.1$ hours vs. $26.1$ hours), while achieving higher performance (see Fig.~\ref{fig:fig4}). To have a more standardised comparison, we finetuned all the models on PTB-XL for the same number of training steps ($30$ epochs and $96$ batch size to give $5430$ training steps) and found that xECG required $\sim10\times$ less time ($7$ min vs. $67$ min) and $\sim7\times$ less memory ($4.2$ GB vs. $28.4$ GB) compared to the second-best method ST-MEM. Note that as ECGFounder is a relatively small CNN ($30.7$M parameters) compared to the other foundation models (ST-MEM $85.2$M, JEPA $85.4$M, xECG $57.0$M, SimDINOv2 Transformer $85.3$M) its runtime and memory usage is similar to xECG. Additionally, although the SimDINOv2 Transformer has a similar parameter count to the other transformer models, its temporal patching process produces less tokens from the same length input signal and so it uses less memory and trains faster.

\section*{Discussion}\label{sec3}

BenchECG enables systematic and rigorous comparison of ECG foundation models across a wide range of tasks. This provides a reliable way to measure progress and encourage reproducibility and higher quality research in the field of cardiac signal analysis. Notably, the benchmark can also serve as a practical decision-making tool for model selection. For a novel cardiac application, the optimal foundation model can be selected by consulting its performance on a similar task within BenchECG. Using this benchmark, xECG achieved the highest overall score, with strong and consistent performance across diverse signals, patient populations, and tasks.

A key result is that xLSTM-based architectures performed particularly well on long-context tasks, such as sleep-apnea classification in overnight recordings and arrhythmia detection in $30$-minute ambulatory ECGs. These findings support the view that recurrent architectures are advantageous when modelling extended physiological signals, where dependencies span minutes rather than seconds. The superiority of xECG in linear probing on these tasks suggests that its pretrained features are inherently better aligned with long-term temporal dependencies, rather than requiring task-specific finetuning.

On short 12-lead ECGs, which are the most widely studied setting~\citep{han2024foundationmodelselectrocardiogramreview}, differences between models after finetuning were small. However, larger gaps emerged under linear probing, indicating that some representations were more transferable without extensive adaptation. Here, finetuning can mask representational differences by overwriting pretrained features, whereas linear probing provides a measure of the intrinsic quality of learnt representations. Prior work evaluating models only on PTB-XL and similar tasks~\cite{Na2024ST-MEM, Kim20204ECGJEPA} may not be able to adequately measure the differences between architectures, highlighting the importance of evaluating a broader range of tasks.

Cross-population tasks exposed additional distinctions. Although ST-MEM achieved the lowest error in age estimation, xECG exhibited smaller generalisation gaps across datasets from Brazil, the USA, Germany, and China. Survival modelling also demonstrated that while concordance indices dropped across populations, when adjusting xECG risk predictions for age and sex, hazard ratios remained predictive, suggesting the prognostic features learnt during training are at least partially transferable across populations. This supports the translation potential of ECG foundation models, but also highlights the need for evaluation across heterogeneous patient cohorts.

Computational comparisons underscore the importance not just of architecture choice (e.g. CNN, transformer, or xLSTM) but also architectural details. Overall, the CNN and xLSTM based models are more computationally efficient than their transformer counterparts. However, when the patching is performed purely temporally (SimDINOv2 Transformer) rather than across leads and time (ECG-JEPA and ST-MEM), the number of tokens, and hence the computational cost, is greatly reduced. Even so, an advantage of recurrent backbones is they provide linear scaling with input length, enabling efficient training and evaluation on long recordings, where transformer-based methods become prohibitive. 
This computational efficiency is likely to be important for deploying foundation models in real-world clinical settings, where resources can be constrained and continuous Holter monitoring and ICU measurements generate extremely long signals.

Different architectures impose different priors. When designing a foundation model for varied tasks these priors should not be too strongly linked to one task in particular or the model may not generalise well. This is the case with the CNN-based ECGFounder model. The model performs well on classification tasks, but the CNN-based architecture aggregates contextual information at the expense of temporal resolution, making it unsuitable for detection tasks. This is evident in the R-peak detection tasks where ECGFounder performs poorly. ECGFounder's results on BenchECG suggest that while CNN-based foundation models can perform well on classification and regression tasks using 12-lead $10$s ECGs, they do not adapt well to long-context tasks or tasks requiring fine temporal information.

Despite its breadth, BenchECG has several limitations. Not all models can be evaluated. If a model has been pretrained on any of the evaluation datasets, it cannot be fairly compared to other methods. Relatedly, like any fixed benchmark, BenchECG risks becoming a target for overfitting as future methods are tuned specifically to its constituent datasets. Although it covers diverse tasks, it remains constrained by the availability of public datasets, which may not capture the full variability of real-world ECG practice. Additionally, there are substantial differences in the pretraining datasets of the models compared in this paper. ECGFounder used HEEDB~\citep{Koscova2024heedb} ($\sim10$M ECGs), xECG used CODE, INCART and Chapman \& Ningbo ($\sim 8$M ECGs), whereas ST-MEM and ECG-JEPA used CODE-15\%, Chapman \& Ningbo ($\sim 400,000$ ECGs, but these works discarded signals shorter than $10$s, giving $\sim 200,000$ ECGs). This limits claims on which pretraining methodology is most suitable for ECG feature learning and future work should explore direct comparisons of pretraining methods using the same pretraining data. Regardless, scaling self-supervised training further and pretraining on a wider variety of ECG signals is likely to be essential, as seen in NLP and vision foundation models \citep{kaplan2020scaling, zhai2022scaling}. Finally, the evaluation focuses on technical metrics and does not directly assess clinical utility or safety, which would require prospective validation in patient care settings. Benchmarking is an important step for model development and comparisons, but should not be seen as the end of the evaluation process.

Overall, the results show that while several foundation models achieve strong performance on conventional ECG benchmarks, xECG provides superior generalisation, flexibility, and computational efficiency. 

\section*{Methods}\label{sec4}

\subsection*{BenchECG datasets and tasks}

BenchECG includes a total of eight public datasets ($7$ for evaluation and $1$ for training) and $10$ tasks. In the following, we describe each task, including the data splits used for training, validation and testing, and how each task is incorporated into the BenchECG score.

\textbf{PTB-XL.} The PTB-XL dataset (Version: 1.0.3) \cite{wagner2020ptb} comprises a collection of $21,799$, $10$ second, $500$ Hz, clinical 12-lead ECGs from $18,869$ patients in Germany. The dataset includes a broad range of diagnostic classes including a large number of healthy records. For our evaluation, we focused on the multilabel classification of five diagnostic superclasses: Normal ECG (NORM), Myocardial Infarction (MI), ST/T Change (STTC), Conduction Disturbance (CD) and Hypertrophy (HYP).
The diagnostic annotations are inherently multilabel, as patients can present with multiple concurrent conditions. We emphasise that we assess performance solely in the multilabel setting as opposed to many other works \cite{Na2024ST-MEM, Kim20204ECGJEPA}, which used this dataset in a simplified fashion, where ECGs with more than one label were not considered. We adopt the official stratified splits \citep{wagner2020ptb}, using fold $10$ for held-out testing and fold $9$ for validation.

\textbf{CPSC2018.} The CPSC2018 dataset \cite{liu2018CPSC} comprises $9,831$ 12-lead-ECGs from $9,458$ patients, recorded in eleven distinct hospitals in China, with lengths ranging from $7$ to $60$ seconds and sampled at $500$ Hz. We included this dataset as it complements the PTB-XL dataset in several aspects: (1) different set of diagnostic multilabels, (2) different patient population, and (3) varying signal length. The BenchECG task using this dataset is a multilabel classification problem including nine label categories: normal, atrial fibrillation (AF), left bundle branch block (LBBB), right bundle branch block (RBBB), first-degree atrioventricular block (I-AVB); premature atrial contraction (PAC), premature ventricular contraction (PVC), ST-segment elevation (STE), and ST-segment depression (STD). The test set for CPSC2018 is private, so we use fold $6$ as validation and fold $7$ as test.

\textbf{MIT-BIH.} To assess performance on a fine-grained (heartbeat-level), multilabel classification task, we used the MIT-BIH Arrhythmia Database (Version: 1.0.0)~\cite{moody2001impact}, from the USA. This dataset consists of $48$, half-hour, two-lead ambulatory ECG recordings from $47$ subjects, digitised at $360$ Hz. Each heartbeat is annotated into one of five arrhythmia classes according to the AAMI standard [citation]: Normal (N), Supraventricular Ectopic (SVE), Ventricular Ectopic (VE), Fusion (F), and Unknown (Q). This dataset is challenging due to its limited size, highly imbalanced classes, and the need for beat-level classification. Following standard practice \cite{de2004automatic}, we use the intra-patient DS1/DS2 split for training and testing, respectively. The specific patient IDs for the split DS1 are: $101$, $106$, $108$, $109$, $112$, $114$, $115$, $116$, $118$, $119$, $122$, $124$, $201$, $203$, $205$, $207$, $208$, $209$, $215$, $220$, $223$, $230$: and in the DS2 split: $100$, $103$, $105$, $111$, $113$, $117$, $121$, $123$, $200$, $202$, $210$, $212$, $213$, $214$, $219$, $221$, $222$, $228$, $231$, $232$, $233$, $234$. As in previous work, we excude four patients with pacemakers. Split DS1 is further divided in train and validation sets keeping patients $114$ and $203$ for validation. 

\textbf{R-peak.} To evaluate R-peak detection performance we also used the MIT-BIH dataset, using the R-peak annotation locations as ground truth for this task. The same train/val/test splitting is used.

\textbf{Sleep Apnea.} To evaluate a model's ability to handle long-duration signals and tasks, we include the PhysioNet Apnea-ECG Database (Version: 1.0.0)~\cite{penzel2000apnea}. This dataset contains $70$ single-lead ECG recordings collected in Germany, each approximately $7$ to $10$ hours in duration, with sleep apnea annotations (presence/absence of sleep apnea) at a one-minute resolution. Each participant has a single ECG. The benchmark task is to segment the signal into presence or absence of sleep apnea at a one-minute resolution. The test set is comprised of the filenames beginning with `x'. The rest of the patients are divided in a random split where 20\% of subjects are chosen to be in the validation set and the rest is used for training.

\textbf{PPG AF.} In order to evaluate foundation models on an atypical and out of distribution scenario we include the DeepBeat PPG Dataset \cite{torres-sotoMultitaskDeepLearning2020}. It provides photoplethysmography (PPG) signals annotated with atrial fibrillation (AF) labels. PPG signals are recorded from wearable devices rather than traditional ECG equipment, presenting unique challenges related to noise artifacts and signal quality that are characteristic of wrist-worn sensors.
The dataset comprises over $500,000$ labelled PPG signals from more than $100$ individuals, collected from three distinct sources to ensure diversity and robustness:
\begin{itemize}
    \item Cardioversion Cohort: $107$ patients undergoing elective cardioversion procedures at Stanford University, all presenting with atrial fibrillation.
    \item Exercise Stress Test Cohort: $41$ participants undergoing elective exercise stress tests, representing normal sinus rhythm cases. 
    \item IEEE Signal Processing Cup $2015$ Dataset: Publicly available data supplementing the Stanford collections to provide out-of-institution examples and enhance generalisability.
\end{itemize}
We used the official train, validation and test splits \cite{torres-sotoMultitaskDeepLearning2020}. 

\textbf{Exercise.} The ECG in High Intensity Exercise Dataset (Exercise-ECG) consist of ECGs extracted in different times of a maximal exercise test. In particular, for each patient the first segment was extracted $30$s before a heavy intensity effort and the second $60$s after. The third $30$s before $VO_2$ max (highly severe intensity up to exhaustion). The fourth at the moment of exhaustion (centred on the VO2 max measurement); the fifth $60$s post-exercise, i.e. during the recovery after exhaustion.
This is a very small dataset with only $20$ participants and $5$ recordings each, for a total of $100$ ECGs, each $20$s long. The dataset includes R-peak annotations which we use to train for R-peak detection. We used patients $10$-$13$ for validation and $13$-$20$ for testing. 

\textbf{Age.} A regression task where models are trained to predict the age of a patient from their ECG. We used the CODE-15\% dataset for training and validation while the entire PTB-XL, CPSC2018 and MIMIC-IV-ECG (Version: 1.0)~\cite{gow2023mimic} datasets for testing in order to assess the performance of the models across different populations (Europe, China and USA, respectively). Ages range from $17$-$100$ in the training set and $1$-$101$ in the testing sets. For the PTB-XL dataset, patients older than $90$ years have a recorded age of $300$ years for privacy reasons, hence we excluded these patients. In the CPSC2018 dataset we did not considered patients with invalid age (e.g., -1 or not a numbers).

\textbf{Mortality.} In order to evaluate foundation models on survival analysis, we again used the CODE-15\% dataset for training and validation and the MIMIC-IV-ECG dataset for testing. CODE-15\% includes follow-up time in years for all ECGs and a binary label indicating mortality. For MIMIC-IV-ECG, we extracted the mortality data from the admission file of the original MIMIC-IV dataset (Version: 3.1)~\cite{johnson2023bMIMICIV}. The models are trained to minimize the Cox proportional hazards partial likelihood function \cite{belloDeepLearningCardiac2019}:
\begin{equation}
    \mathcal{L}_\mathrm{cox} = \sum_{i=1}^n\delta_i\Bigl\{\phi(x_i) - \text{ln}\sum_{j\in R(t_i)}e^{\phi(x_j)}\Bigl\}
\end{equation}
where $\delta_i$ is a boolean indicator of the subject $i$'s status (mortality) and $\phi(x_i)$ is the scalar output of the model, representing the survival prediction (specifically, the natural logarithm of the hazard ratio) and $R(t_i)$ is the subset of patients not censored at the time subject $i$ died or became censored $(\{ j : t_j > t_i\})$.

\textbf{Blood test.} To assess the capability of foundation models to extract patient information normally collected separately from an ECG, we predict (ab)normal blood test results for patients in the MIMIC-IV-ECG dataset. We extract the labels from the original MIMIC-IV dataset~\cite{johnson2023bMIMICIV}. Previous work analysing blood marker prediction from MIMIC-IV-ECG~\cite{miguellopezalcarazCardioLabLaboratoryValues2024} found many markers are too challenging to predict from ECG signals. In the benchmark we evaluate only those markers which achieved an AUROC above $0.7$ in this prior work~\cite{miguellopezalcarazCardioLabLaboratoryValues2024}. These markers are: Albumin; Hemoglobin; NTproBNP; Acetaminophen; Hematocrit; PT; Red Blood Cells; 25-OH Vitamin D; RDW-SD; INR(PT); Urea Nitrogen; Monocytes; Acetaminophen; Absolute Basophil Count; Urea Nitrogen; C-Reactive Protein; Cholesterol, HDL; Bilirubin, Direct; Creatinine; Sedimentation Rate; pO2; Osmolality, Measured; Bicarbonate. We treat the task as a multi-task, multi-class setting. Where the model predicts whether the result is below, inside or above the reference values. 
Reference values considered for the classification tasks are chosen to be the median values of the reference values, following the motivation of the work cited before.
For the test split we used patients in fold 19, for the validation, fold 18, and the rest for training. These folds come from the diagnostic labels in the MIMIC-IV-ECG-Ext-ICD external dataset (Version: 1.0.1)~\cite{strodthoff2024mimic}.

\subsection*{BenchECG Metrics} \label{sec: Methods - Metrics}
In order to define a BenchECG score, each task needs to be evaluated using a metric normalised between $0$-$1$, with $1$ indicating perfect performance. In this section we justify our choice of metric for each task.

For classification tasks we use area under the receiver operator characteristic curve (AUROC). AUROC is a suitable metric as it is not affected by prevalence and it represents a model's predictive power across all operating points. An AUROC of $1$ indicates perfect classification and an AUROC of $0.5$ is equivalent to a random model. The one exception to using AUROC is for MIT-BIH, where we use F1 score. This is because F1 score is typically reported in the literature \cite{xia2023transformer,sellami2019robust, li2022inter, marinho2019novel, li2019automated}, unlike AUROC, and so allows for better comparisons to previous work. 

For R-peak detection, we also use F1. However, more detail is required, as how close does a prediction need to be for it to count as correct? A recent review on R-peak detection \cite{aliReliableECGAnalysis2024} suggests using a tolerance window of $150$ms, but, as can be seen from our results, this leads to saturated performance with many methods being near-perfect. Instead, we use a stricter $20$ms total tolerance window to push the boundaries of this task and have a more useful comparison across methods.

For age regression, we use the symmetric mean absolute percentage error (SMAPE) wich is bounded between $0$ and $1$:
\begin{equation}
    \text{SMAPE} = \frac{1}{n}\sum_{i=1}^{n}\frac{|x_i - y_i|}{|x_i| + |y_i|}
\end{equation}
where $x_i$ is the model prediction and $y_i$ is the ground truth. An SMAPE of $0$ indicates no error so, for our BenchECG score, we use $1 -$ SMAPE.

For the survival analysis task, we use concordance index. Concordance index represents the probability that two randomly selected subjects will maintain the same relative order of their observed event (death in this case) as predicted by the model. A higher C-index, closer to $1$, indicates better predictive performance, while a value of $0.5$ suggests the model is no better than random guessing, similar to AUROC.

\subsection*{Model selection and data preprocessing} \label{sec: Model Selection}

Our primary baselines are ST-MEM~\cite{Na2024ST-MEM}, ECG-JEPA~\cite{Kim20204ECGJEPA} and ECG founder~\cite{Li2025Founder}. To ensure a fair and direct comparison, we use the official pretrained weights provided by the authors and re-evaluate them within our standardised evaluation framework. For each model we applied the preprocessing strategies as detailed in their respective original works~\cite{Na2024ST-MEM, Kim20204ECGJEPA, Li2025Founder}. Below, we briefly describe each model.

\textbf{ST-MEM.}  A foundation model based on the vision transformer (ViT-Base) architecture~\cite{dosovitskiy2021an}. It is pretrained in a self-supervised manner following a masked auto-encoder paradigm (MAE)~\cite{He_2022_CVPR}. The authors introduce spatio-temporal patchifying, where the signals are tokenised across both time and leads.

\textbf{ECG-JEPA.} A foundation model based on the vision transformer (ViT-Base) architecture~\cite{dosovitskiy2021an}. It is pretrained in a self-supervised manner following JEPA~\citep{assran2023self}. The authors introduce a custom cross pattern attention~\cite{Kim20204ECGJEPA} which replaces standard attention. As in ST-MEM, signals are tokenised across both time and leads.

\textbf{ECGFounder.} This foundation model is based on convolutional neural network (CNN) and was trained with supervision on a large scale labelled dataset of over 10 million ECGs~\cite{Li2025Founder}. Here the model accept the full signal in input and outputs a single feature map. 

\textbf{SimDINOv2 Transformer.} This is an ablation study for xECG, where we use a vision transformer (ViT-Base) architecture~\cite{dosovitskiy2021an} in place of the xLSTM, but apply the same simDINOv2 pretraining and the same patching strategy. 

\textbf{xLSTM Supervised.} A further ablation of our xECG model where we maintain the same exact architecture but we do not pre-train it.

Together, these models span different architectural approaches currently used for ECG analysis, including transformer-based, convolutional, and recurrent networks.

\subsection*{xECG Architecture}

The core of our xECG is an encoder composed of a stack of nine alternating sLSTM and mLSTM blocks (s, s, m, m, s, s, m, m, s) processing sequences of ECG patches bidirectionally. A high-level overview of our architecture is shown in Figure~\hyperref[fig:fig1]{1b}.

Let an input ECG signal be denoted by $s \in \mathbb{R}^{L\times C}$, where $L$ is the signal length and $C$ is the number of leads (channels). Given a model sampling frequency $f_m \in \mathbb{R}_{\geq 0}$ and an input signal frequency $f_s \in \mathbb{R}_{\geq 0}$, the signal is first resampled to $f_m$. 
We intentionally omit any other kind of pre-processing steps, e.g., normalisation or extensive filtering. Thus, the model directly learns  from the raw signal, allowing it to capture any potentially relevant information present in the signal.

The resampled signal is then divided along only the temporal dimension into a sequence of $N$ non-overlapping patches $P = (p_e, p_1, p_2,...,p_N)$, where each patch has a size of $P_s$. The signal is truncated to ensure its length is a multiple of $P_s$. Each patch is flattened and transformed via a linear projection into the model's embedding space, resulting in a sequence of patch embeddings $s_e = (e_1, e_2, ..., e_N)$, where each $e_i \in \mathbb{R}^E$ and $E$ is the model's embedding dimension.

The sequence of patch embeddings $s_e$ is processed by the xLSTM encoder. The encoder consists of stacked pairs of sLSTM and mLSTM blocks. To achieve bidirectionality, we adopt a layer-wise alternating strategy. Each pair of blocks processes the sequence in opposing directions: one block processes the sequence from start to finish, while the second processes a reversed version.

After passing through the final encoder layer, we obtain a sequence of rich patch representations $s_r$. These patches can be used in different ways: pooled to produce a single, fixed-size vector for downstream classification or as they are for segmentation or detection tasks. During pretraining these representations are aggregated using an attention pooling layer \cite{bolya2025perceptionencoderbestvisual}.

\subsection*{Self-supervised pretraining}

The adopted SimDINOv2 \cite{wu2025simplifying} pretraining strategy is based on a teacher-student framework where the student network learns by matching the output of the teacher network. The student parameters $\theta_t$ are updated via standard backpropagation. The teacher parameters $\theta_t$ are not trained directly but are instead an exponential moving average (EMA) of the student's weights. At each training iteration, the teacher is updated using the following rule:

\begin{equation}
    \theta_t \leftarrow \lambda_t\theta_t + (1 - \lambda_t) \theta_s
\end{equation}
where $\lambda_t$ is the momentum parameter scheduled to increase during training. This schedule encourages the teacher to follow the student's progress more closely at the beginning and then stabilize as training progresses. The schedule is defined as:

\begin{equation}
\lambda_t \leftarrow \lambda_\mathrm{base} + (t/N)(1-\lambda_\mathrm{base})
\end{equation}
Here, $\lambda_\mathrm{base}$ is the initial momentum value, $t$ is the current training iteration and $N$ is the total number of training iterations.

A key component of the DINO family of algorithms is the use of multi-crop augmentations, where the model learns to associate different views of the same sample. We adapt this concept from the image domain to time-series ECG data. 
For each ECG signal in a batch, we generate a set of augmented ``views" by creating random sub-sequences of varying lengths:
\begin{itemize}
    \item \textbf{Global Views}: Two long, overlapping sub-sequences (e.g., $6$-$10$ seconds long). 
    \item \textbf{Local Views}: Four shorter sub-sequences (e.g., $1$-$3$ seconds long). 
\end{itemize}
In Fig.~\hyperref[fig:fig1]{1c} an example of different views of the same sample is shown.

During training, the teacher network processes only the global views. The student network, instead, process all views (both global and local). The training objective will force the student's output for every view to match the teacher's output for the corresponding global view. 
Because in the pretraining we might have multiple ECGs for each patient, we consider different samples of the same patient as a single source where to get different views. This means that if a patient has two ECGs, a global (or local) view is selected to be a slice of one of the two samples, selected randomly. At loss level we treat then these two views coming from two different samples (but same patient) as if they belong to the same original signal.
This loss objective of SimDINOv2 consists of three distinct components: 
\begin{itemize}
\item \textbf{Patch-level objective ($\mathcal{L}_\mathrm{patch}$)}:  For each global view provided to the student network, a random subset of its patch embeddings are replaced with a shared, learnable $[\textsf{MASK}]$ token. The teacher network receives the same global view but without any masking. The student is then asked to reconstruct the original, unmasked patch representations from the teacher. The loss is computed as the Mean Squared Error (MSE) between the student's normalised output for the masked positions and the teacher's corresponding outputs. Formally, let $\hat{s}_t$ be the sequence of $l_2$ normalised patch representations from the teacher for an unmasked view, and let $\hat{s}_s$ be the $l_2$ normalised student's output for the masked view. If $M$ is the set of indices for the masked patches, the loss is:
\begin{equation}
    \mathcal{L}_\mathrm{patch} = \frac{1}{|M|} \sum_{i \in M} \left\| \hat{s}_{s,i} - \hat{s}_{t,i} \right\|_2^2,
\end{equation}
with
\begin{equation}
    \hat{s}_{s, i} = \frac{s_{s, i}}{\left\|s_{s, i}\right\|_2},\,  \hat{s}_{t, i} = \frac{s_{t, i}}{\left\|s_{t, i}\right\|_2}
\end{equation}
\item \textbf{Sample-level objective ($\mathcal{L}_\mathrm{view}$):} The student network is trained to produce representations for all views (both global and local) of the same sample to be closer to the respective teacher's global views representations. Specifically, the loss is calculated between the student's representation for one view and the teacher's representation for a different global view from the same original signal. This is performed across all pairs of student-teacher views. 
Let $K$ be the total amount of global and local views and $Z_t = \{z_{t,1}, ..., z_{t,G}\}$ be the set of pooled representations of the global views processed by the teacher, where $G<K$ is the number of global views.  Let $Z_s = \{z_{s,1}, ..., z_{s,K}\}$ be the set of representations of all the $K$ views presented to the student, the loss is defined as:
\begin{equation}
\mathcal{L}_\mathrm{view} = \sum_{i=1}^G \sum_{\substack{j=1\\ j\neq i }}^{K} \left\| \hat{z}_{t,i} - \hat{z}_{s,j} \right\|_2^2,
\end{equation}
with
\begin{equation}
    \hat{z}_{s, i} = \frac{z_{s, i}}{\left\|z_{s, i}\right\|_2},\,\hat{z}_{t, i} = \frac{z_{t, i}}{\left\|z_{t, i}\right\|_2}
\end{equation}
\item \textbf{Coding rate regularizer ($\mathcal{L}_\mathrm{cr}$)}: 
This regularizer operates on the covariance matrix of the $l_2$ normalised feature vectors $\hat{z}_s$ produced by the student network. 
Formally, the regularization term to be minimised is:
\begin{equation}
\mathcal{L}_\mathrm{cr} = - \gamma R_{\epsilon}(Cov[\hat{z}_s])
\end{equation}
where $\gamma$ is a scalar hyperparameter that controls the strength of the regularization defined by:
\begin{equation}
    \gamma = \epsilon\sqrt{B/(E\min\{E, B\})} 
\end{equation}
being $\epsilon$ an hyperparameter, $E$ the embedding size and $B$ the batch size.
$Cov[\hat{z}_s]$ is the sample covariance matrix of the student's normalised feature vectors within a batch, and $R_\epsilon$ is the coding rate function defined as:
\begin{equation}
R_{\epsilon}(\Gamma) = \frac{1}{2} \log \det \left( \mathbf{I} + \frac{E}{\epsilon} \Gamma \right)
\end{equation}
Here, $\Gamma$ is the covariance matrix and $I$ is the identity matrix. 
\end{itemize}
The final composite loss $\mathcal{L}$ is then the sum of these terms:
\begin{equation}
\mathcal{L}_\mathrm{total} = \mathcal{L} _\mathrm{patch} + \mathcal{L} _\mathrm{view} + \mathcal{L}_\mathrm{cr}
\end{equation}

Adapting the multi-crop strategy from the image domain to ECG signals requires a set of carefully designed augmentations:

\begin{itemize}
\item \textbf{Random lead dropout}\cite{oh2022lead}: Every lead is zeroed out with a probability $p_\mathrm{drop}$. This do not apply to lead $\textsf{II}$ to prevent all leads are zeroed out. This forces the model to learn redundant information present across the 12 leads and reconstruct a complete cardiac picture from partial data.

\item \textbf{Low-frequency component swap:} To teach invariance to baseline wander caused by patient movement or respiration, we isolate and swap low-frequency components across signals in a batch. First, we apply a low-pass filter (a Butterworth filter with a cutoff at 0.5 Hz) to extract the low-frequency baseline from each signal. These baseline signals are then randomly shuffled and added back to the corresponding high-frequency components of different signals in the batch.

\item \textbf{Multiplicative gaussian jitter:} To simulate realistic sensor noise that often scales with signal magnitude, we apply a multiplicative form of Gaussian jitter. With a probability $p_\mathrm{jitter}$, the augmented signal $s'(t)$ is generated from the original signal $s(t)$ according to the formula:
\begin{equation}
    s'(t) = s(t) \cdot \left[1 + A \cdot n(t)\right]
\end{equation}
where $A$ is a scalar ``amplitude" hyperparameter, and $n(t)$ is noise sampled from a standard normal distribution $\mathcal{N}(0, \sigma^2)$. 
\item \textbf{Random amplitude scaling:} To account for variations in overall signal strength due to factors like electrode-skin impedance, the entire signal is scaled by a single random scalar $\alpha$. With a probability $p_\mathrm{scale}$, the scalar $\alpha$ is drawn from a uniform distribution:
\begin{equation}
    \alpha \sim \mathcal{U}\left(1 - \frac{R}{2}, 1 + \frac{R}{2}\right)
\end{equation}
where $R$ is the amplitude range hyperparameter. The augmented signal is then $s'(t) = \alpha \cdot s(t)$.
\end{itemize}

This pretraining strategy used a large-scale corpus of ECG data aggregated from several publicly available sources:
\begin{itemize}
    \item \textbf{CODE}~\cite{ribeiro2019tele}: This extensive dataset comprises 2,322,513 12-lead ECG recordings from 1,676,384 patients. The signals have durations ranging from 7 to 10 seconds and were recorded at various sampling rates between 300 and 600 Hz.
    \item \textbf{Chapman}~\cite{zheng202012} and \textbf{Ningbo}~\cite{zheng2020optimal}: Together, these datasets contribute 45,152 10-second, 12-lead ECGs sampled at 500 Hz.
    \item \textbf{Incart}~\cite{goldberger2000physiobank}: This dataset consists of 75 12-lead Holter recordings, each 30 minutes in duration and sampled at 257 Hz, with each recording from a different patient. These long-duration signals provide a source for extracting multiple, diverse segments from a single continuous recording.
\end{itemize}
To ensure a minimum standard of quality and consistency across the aggregated datasets, a simple yet effective preprocessing pipeline was applied to each ECG signal. We filter out corrupted or invalid recordings. We excluded any signals that met one of the following criteria:
\begin{enumerate}
    \item The signal contained missing (NaN) values.
    \item The signal was completely composed by zeroes.
    \item The signal exhibited excessive noise or artifacts, identified by a variance greater than 10 combined with an absolute amplitude exceeding 15 mV.
\end{enumerate}
Then, all signals were uniformly resampled to a target frequency of 100 Hz. This step serves a dual purpose: it standardises the temporal resolution and acts as a low-pass filter, removing high-frequency noise while preserving the essential morphological features of the ECG waveform. Lower sampling rate reduces computational complexity without compromising downstream task performance \cite{mehari2022advancing}.

\subsection*{Data availability}

All datasets used in BenchECG are publicly available:
\begin{itemize}
    \item PTB-XL: freely available at \url{https://physionet.org/content/ptb-xl/1.0.3/}.
    \item CPSC2018: freely available at \url{https://physionet.org/content/challenge-2020/1.0.2/training/cpsc_2018/}.
    \item MIT-BIH Arrhythmia Database: freely available at \url{https://www.physionet.org/content/mitdb/1.0.0/}.
    \item ECG in High Intensity Exercise Dataset: freely available at \url{https://zenodo.org/records/5727800}.
    \item Sleep Apnea-ECG Database: freely available at \url{https://www.physionet.org/content/apnea-ecg/1.0.0/}.
    \item DeepBeat PPG: available at \url{https://www.synapse.org/Synapse:syn21985690/wiki/}.
    \item MIMIC-IV-ECG (v1.0): available at \url{https://physionet.org/content/mimic-iv-ecg/1.0/}.
    \item MIMIC-IV-ECG-Ext-ICD (v1.0.1): available at \url{https://physionet.org/content/mimic-iv-ecg-ext-icd-labels/1.0.1/} (requires PhysioNet credential access and training completion).
    \item MIMIC-IV (v3.1): available at \url{https://physionet.org/content/mimiciv/3.1/} (requires PhysioNet credential access and training completion).
    \item CODE-15: freely available at \url{https://zenodo.org/records/4916206}.
\end{itemize}

For pre-training we use the CODE dataset. For access, contact the authors of~\cite{ribeiro2019tele}. Additionally we used INCART (available at url{https://physionet.org/content/incartdb/1.0.0/}) and Chapman \& Ningbo (available at \url{https://physionet.org/content/ecg-arrhythmia/1.0.0/}).

\subsection*{Code availability}
We release xECG weights and the BenchECG code in the following repository: \url{https://github.com/dlaskalab/bench-xecg/}.

\newpage
\clearpage

\bibliography{sn-bibliography}

\newpage
\clearpage
\section*{Supplementary Information}

\subsection*{Individual BenchECG Results}

In this section, we report the individual results of each model for each task in BenchECG. Each model is trained $5$ times and we report the mean$\pm$ standard deviation (SD).  Pairwise differences between models were assessed using two-sided Welch's t-tests across independent runs, with significance defined as $p<0.05$. Bold results indicate best performance across models and underlined indicates 2nd best.

\begin{table*}[h!]
\centering
{\footnotesize
    \begin{tabular}{l|cccc}
    \toprule
    \multirow{2}{*}{\textbf{Methods}} & \multicolumn{4}{c}{\textbf{PTB-XL}} \\
    & Accuracy& F1 Score& AUROC& AUPRC \\
    \midrule
    \multicolumn{5}{c}{\textbf{Supervised}} \\
    \midrule
     xLSTM-ECG \cite{kang2025xlstm} &  0.875 & - & 0.913 & -\\
     ASTLNet \cite{10080902}  &  0.629 & - & 0.913 & -\\
     ECGNet \cite{8438739} & 0.873 & - & 0.901 & -\\
   \textbf{Supervised xLSTM} & 0.860 $\pm$ 0.005 & 0.667 $\pm$ 0.007 & 0.891 $\pm$ 0.004 & 0.748 $\pm$ 0.006 \\
    \midrule
    \multicolumn{5}{c}{\textbf{Linear Probing}} \\
    \midrule
   ST-MEM & 0.847 $\pm$ 0.001 & 0.564 $\pm$ 0.002 & 0.867 $\pm$ 0.001 & 0.693 $\pm$ 0.001 \\
   ECG-JEPA & 0.864 $\pm$ 0.001 & \textbf{0.712 $\pm$ 0.002} & 0.902 $\pm$ 0.000 & 0.769 $\pm$ 0.000 \\
   ECGFounder & \textbf{0.883 $\pm$ 0.001} & \underline{0.702 $\pm$ 0.002} & \textbf{0.917 $\pm$ 0.000} & \textbf{0.799 $\pm$ 0.000} \\
   \textbf{SimDINOv2 Transformer} & 0.875 $\pm$ 0.001 & \underline{0.703 $\pm$ 0.002} & 0.914 $\pm$ 0.000 & 0.782 $\pm$ 0.001 \\
   \textbf{xECG} & \underline{0.876 $\pm$ 0.001} & 0.690 $\pm$ 0.001 & \underline{0.915 $\pm$ 0.000} & \underline{0.788 $\pm$ 0.001} \\
    \midrule
    \multicolumn{5}{c}{\textbf{Finetuning}} \\
    \midrule
   ST-MEM & \textbf{0.892 $\pm$ 0.001} & \textbf{0.739 $\pm$ 0.005} & \textbf{0.930 $\pm$ 0.001} & \textbf{0.822 $\pm$ 0.002} \\
   ECG-JEPA & 0.884 $\pm$ 0.001 & 0.733 $\pm$ 0.003 & 0.923 $\pm$ 0.000 & 0.809 $\pm$ 0.000 \\
   ECGFounder & \textbf{0.891 $\pm$ 0.001} & \textbf{0.734 $\pm$ 0.002} & \textbf{0.930 $\pm$ 0.000} & \textbf{0.822 $\pm$ 0.001} \\
   \textbf{SimDINOv2 Transformer} & 0.880 $\pm$ 0.002 & 0.710 $\pm$ 0.017 & 0.925 $\pm$ 0.001 & 0.808 $\pm$ 0.001 \\
   \textbf{xECG} & 0.884 $\pm$ 0.002 & 0.718 $\pm$ 0.011 & 0.928 $\pm$ 0.001 & 0.816 $\pm$ 0.003 \\
    \bottomrule
    \end{tabular}
}
\vspace{0.5em}
\caption{\textbf{PTB-XL.} Individual model performance for the PTB-XL task. Each model is trained $5$ times and we report the mean $\pm$ standard deviation.  Pairwise differences between models were assessed using two-sided Welch's t-tests across independent runs, with significance defined as $p<0.05$. Bold results indicate best performance across models and underlined indicates 2nd best.}
\label{tab:ptb_xl_results}
\end{table*}

\begin{table*}[h!]
\centering
{\footnotesize
    \begin{tabular}{l|cccc}
    \toprule
    \multirow{2}{*}{\textbf{Methods}} & \multicolumn{4}{c}{\textbf{CPSC2018}} \\
    & Accuracy& F1 Score& AUROC& AUPRC \\
    \midrule
    \multicolumn{5}{c}{\textbf{Supervised}} \\
    \midrule
   \textbf{Supervised xLSTM} & 0.952 $\pm$ 0.001 & 0.719 $\pm$ 0.008 & 0.951 $\pm$ 0.002 & 0.813 $\pm$ 0.006 \\
    \midrule
    \multicolumn{5}{c}{\textbf{Linear Probing}} \\
    \midrule
   ST-MEM & 0.935 $\pm$ 0.000 & 0.455 $\pm$ 0.005 & 0.922 $\pm$ 0.001 & 0.722 $\pm$ 0.004 \\
   ECG-JEPA & 0.889 $\pm$ 0.002 & 0.613 $\pm$ 0.003 & 0.961 $\pm$ 0.000 & \underline{0.830 $\pm$ 0.001} \\
   ECGFounder & \underline{0.958 $\pm$ 0.000} & \underline{0.750 $\pm$ 0.007} & \underline{0.962 $\pm$ 0.000} & \underline{0.830 $\pm$ 0.001} \\
   \textbf{SimDINOv2 Transformer} & 0.953 $\pm$ 0.001 & 0.733 $\pm$ 0.008 & 0.951 $\pm$ 0.001 & 0.787 $\pm$ 0.008 \\
   \textbf{xECG} & \textbf{0.961 $\pm$ 0.002} & \textbf{0.781 $\pm$ 0.011} & \textbf{0.968 $\pm$ 0.001} & \textbf{0.861 $\pm$ 0.004} \\
    \midrule
    \multicolumn{5}{c}{\textbf{Finetuning}} \\
    \midrule
   ST-MEM & \underline{0.963 $\pm$ 0.001} & \underline{0.789 $\pm$ 0.012} & 0.958 $\pm$ 0.003 & 0.835 $\pm$ 0.011 \\
   ECG-JEPA & 0.942 $\pm$ 0.003 & 0.749 $\pm$ 0.009 & 0.965 $\pm$ 0.001 & 0.842 $\pm$ 0.003 \\
   ECGFounder & \underline{0.964 $\pm$ 0.000} & \underline{0.794 $\pm$ 0.005} & \underline{0.969 $\pm$ 0.001} & \underline{0.854 $\pm$ 0.002} \\
   \textbf{SimDINOv2 Transformer} & 0.961 $\pm$ 0.001 & \underline{0.796 $\pm$ 0.008} & \underline{0.965 $\pm$ 0.003} & \underline{0.853 $\pm$ 0.006} \\
   \textbf{xECG} & \textbf{0.968 $\pm$ 0.001} & \textbf{0.822 $\pm$ 0.008} & \textbf{0.981 $\pm$ 0.001} & \textbf{0.888 $\pm$ 0.003} \\
    \bottomrule
    \end{tabular}
}
\vspace{0.5em}
\caption{\textbf{CPSC2018.} Individual model performance for the CPSC2018 task. Each model is trained $5$ times and we report the mean$\pm$ standard deviation.  Pairwise differences between models were assessed using two-sided Welch's t-tests across independent runs, with significance defined as $p<0.05$. Bold results indicate best performance across models and underlined indicates 2nd best.}
\label{tab:cpsc2018_results}
\end{table*}

\begin{table*}[h!]
\centering
{\footnotesize
    \begin{tabular}{l|ccccc}
    \toprule
    \multirow{2}{*}{\textbf{Methods}} & \multicolumn{5}{c}{\textbf{MIT-BIH}} \\
    & Accuracy& Sensitivity& PPV& Specificity& F1 Score \\
    \midrule
    \multicolumn{6}{c}{\textbf{Supervised}} \\
    \midrule
     Xia et al.\cite{xia2023transformer} & 0.976 & 0.510 & 0.563 & 0.930 & 0.529 \\
     Sellami et al. \cite{sellami2019robust} & 0.953 & 0.690 & 0.555 & 0.960 & 0.558 \\
     Li et al.\cite{li2022inter} & 0.889 & 0.521 & 0.568 & 0.947 & 0.533 \\
     Marinho et al. \cite{marinho2019novel} & 0.943 & 0.478 & - & 0.914 & - \\
     Li et al. \cite{li2019automated} & 0.914 & 0.616 & 0.489 & 0.950 & 0.539 \\
   \textbf{Sup. xLSTM} & 0.896 $\pm$ 0.013 & 0.448 $\pm$ 0.028 & 0.416 $\pm$ 0.020 & 0.910 $\pm$ 0.004 & 0.402 $\pm$ 0.014 \\
    \midrule
    \multicolumn{6}{c}{\textbf{Linear Probing}} \\
    \midrule
   ST-MEM & 0.937 $\pm$ 0.003 & 0.424 $\pm$ 0.027 & 0.507 $\pm$ 0.035 & 0.920 $\pm$ 0.003 & 0.436 $\pm$ 0.036 \\
   ECG-JEPA & \underline{0.946 $\pm$ 0.000} & \underline{0.470 $\pm$ 0.003} & \underline{0.574 $\pm$ 0.003} & \underline{0.933 $\pm$ 0.000} & \underline{0.496 $\pm$ 0.003} \\
   ECGFounder & 0.865 $\pm$ 0.012 & 0.210 $\pm$ 0.004 & 0.209 $\pm$ 0.004 & 0.808 $\pm$ 0.003 & 0.206 $\pm$ 0.005 \\
   \textbf{SimDINOv2 Transf.} & 0.931 $\pm$ 0.000 & 0.381 $\pm$ 0.004 & 0.549 $\pm$ 0.012 & 0.924 $\pm$ 0.004 & 0.395 $\pm$ 0.005 \\
   \textbf{xECG} & \textbf{0.972 $\pm$ 0.002} & \textbf{0.686 $\pm$ 0.021} & \textbf{0.669 $\pm$ 0.014} & \textbf{0.976 $\pm$ 0.003} & \textbf{0.674 $\pm$ 0.013} \\
    \midrule
    \multicolumn{6}{c}{\textbf{Finetuning}} \\
    \midrule
   ST-MEM & \underline{0.967 $\pm$ 0.002} & \textbf{0.625 $\pm$ 0.007} & \underline{0.671 $\pm$ 0.013} & \textbf{0.965 $\pm$ 0.003} & \underline{0.644 $\pm$ 0.007} \\
   ECG-JEPA & 0.954 $\pm$ 0.001 & 0.557 $\pm$ 0.006 & 0.629 $\pm$ 0.007 & 0.951 $\pm$ 0.001 & 0.583 $\pm$ 0.007 \\
   ECGFounder & 0.863 $\pm$ 0.008 & 0.211 $\pm$ 0.002 & 0.207 $\pm$ 0.002 & 0.809 $\pm$ 0.002 & 0.207 $\pm$ 0.002 \\
   \textbf{SimDINOv2 Transf.} & 0.896 $\pm$ 0.008 & 0.475 $\pm$ 0.002 & 0.457 $\pm$ 0.005 & 0.905 $\pm$ 0.001 & 0.425 $\pm$ 0.006 \\
   \textbf{xECG} & \textbf{0.976 $\pm$ 0.007} & \textbf{0.651 $\pm$ 0.035} & \textbf{0.721 $\pm$ 0.019} & \textbf{0.973 $\pm$ 0.010} & \textbf{0.677 $\pm$ 0.025} \\
    \bottomrule
    \end{tabular}
}
\vspace{0.5em}
\caption{\textbf{MIT-BIH.} Individual model performance for the MIT-BIH task. Each model is trained $5$ times and we report the mean$\pm$ standard deviation.  Pairwise differences between models were assessed using two-sided Welch's t-tests across independent runs, with significance defined as $p<0.05$. Bold results indicate best performance across models and underlined indicates 2nd best.}
\label{tab:mitbih_results}
\end{table*}

\begin{table*}[h!]
\centering
{\footnotesize
    \begin{tabular}{l|cc}
    \toprule
    \multirow{2}{*}{\textbf{Methods}} & \multicolumn{2}{c}{\textbf{MIT-BIH (R-peak)}} \\
    & F1 Score (150ms)& F1 Score (20ms) \\
    \midrule
    \multicolumn{3}{c}{\textbf{Supervised}} \\
    \midrule
   \textbf{Supervised xLSTM} & 0.980 $\pm$ 0.007 & 0.889 $\pm$ 0.012 \\
    \midrule
    \multicolumn{3}{c}{\textbf{Linear Probing}} \\
    \midrule
   ST-MEM & \underline{0.970 $\pm$ 0.001} & \underline{0.564 $\pm$ 0.002} \\
   ECG-JEPA & \textbf{0.976 $\pm$ 0.000} & 0.444 $\pm$ 0.001 \\
   ECGFounder & 0.073 $\pm$ 0.005 & 0.010 $\pm$ 0.001 \\
   \textbf{SimDINOv2 Transformer} & 0.864 $\pm$ 0.001 & 0.476 $\pm$ 0.000 \\
   \textbf{xECG} & 0.945 $\pm$ 0.005 & \textbf{0.590 $\pm$ 0.011} \\
    \midrule
    \multicolumn{3}{c}{\textbf{Finetuning}} \\
    \midrule
   ST-MEM & \underline{0.994 $\pm$ 0.001} & \textbf{0.937 $\pm$ 0.001} \\
   ECG-JEPA & \textbf{0.996 $\pm$ 0.000} & 0.909 $\pm$ 0.004 \\
   ECGFounder & 0.081 $\pm$ 0.006 & 0.011 $\pm$ 0.001 \\
   \textbf{SimDINOv2 Transformer} & 0.983 $\pm$ 0.001 & 0.900 $\pm$ 0.001 \\
   \textbf{xECG} & \underline{0.995 $\pm$ 0.000} & \underline{0.921 $\pm$ 0.001} \\
    \bottomrule
    \end{tabular}
}
\vspace{0.5em}
\caption{\textbf{R-peak.} Individual model performance for the R-peak detection task (MIT-BIH). Each model is trained $5$ times and we report the mean$\pm$ standard deviation.  Pairwise differences between models were assessed using two-sided Welch's t-tests across independent runs, with significance defined as $p<0.05$. Bold results indicate best performance across models and underlined indicates 2nd best.}
\label{tab:mitbih_r_peak_results}
\end{table*}

\begin{table*}[h!]
\centering
{\footnotesize
    \begin{tabular}{l|cc}
    \toprule
    \multirow{2}{*}{\textbf{Methods}} & \multicolumn{2}{c}{\textbf{Exercise (R-peak)}} \\
    & F1 Score (150ms)& F1 Score (20ms) \\
    \midrule
    \multicolumn{3}{c}{\textbf{Supervised}} \\
    \midrule
   \textbf{Supervised xLSTM} & 0.990 $\pm$ 0.002 & 0.966 $\pm$ 0.002 \\
    \midrule
    \multicolumn{3}{c}{\textbf{Linear Probing}} \\
    \midrule
   ST-MEM & 0.822 $\pm$ 0.018 & \underline{0.627 $\pm$ 0.008} \\
   ECG-JEPA & \underline{0.924 $\pm$ 0.004} & 0.526 $\pm$ 0.007 \\
   ECGFounder & 0.421 $\pm$ 0.016 & 0.083 $\pm$ 0.002 \\
   \textbf{SimDINOv2 Transformer} & 0.603 $\pm$ 0.003 & 0.301 $\pm$ 0.001 \\
   \textbf{xECG} & \textbf{0.948 $\pm$ 0.004} & \textbf{0.892 $\pm$ 0.004} \\
    \midrule
    \multicolumn{3}{c}{\textbf{Finetuning}} \\
    \midrule
   ST-MEM & \underline{0.988 $\pm$ 0.002} & \textbf{0.986 $\pm$ 0.001} \\
   ECG-JEPA & 0.975 $\pm$ 0.002 & 0.859 $\pm$ 0.007 \\
   ECGFounder & 0.423 $\pm$ 0.006 & 0.082 $\pm$ 0.005 \\
   \textbf{SimDINOv2 Transformer} & 0.853 $\pm$ 0.008 & 0.633 $\pm$ 0.013 \\
   \textbf{xECG} & \textbf{0.996 $\pm$ 0.001} & \underline{0.968 $\pm$ 0.002} \\
    \bottomrule
    \end{tabular}
}
\vspace{0.5em}
\caption{\textbf{Exercise.} Individual model performance for the Exercise task (R-peak detection in Exercise-ECG). Each model is trained $5$ times and we report the mean$\pm$ standard deviation.  Pairwise differences between models were assessed using two-sided Welch's t-tests across independent runs, with significance defined as $p<0.05$. Bold results indicate best performance across models and underlined indicates 2nd best.}
\label{tab:exercise_r_peak_results}
\end{table*}

\begin{table*}[h!]
\centering
{\footnotesize
    \begin{tabular}{l|ccc}
    \toprule
    \multirow{2}{*}{\textbf{Methods}} & \multicolumn{3}{c}{\textbf{Physionet Sleep Apnea-ECG}} \\
    & Accuracy& F1 Score& AUROC \\
    \midrule
    \multicolumn{4}{c}{\textbf{Supervised}} \\
    \midrule
     Chang et al.\cite{chang2020sleep} &  0.920 &  0.879 &  0.865 \\
     Shen et al. \cite{shen2021multiscale} &  0.894 & 0.866 & 0.946 \\
     Bernardini et al. \cite{Bernardini2021AIOSASleepApnea} & 0.936 &  0.916 & 0.981 \\
    \textbf{Supervised xLSTM} & 0.778 $\pm$ 0.024 & 0.668 $\pm$ 0.059 & 0.853 $\pm$ 0.022 \\
    \midrule
    \multicolumn{4}{c}{\textbf{Linear Probing}} \\
    \midrule
   ST-MEM & 0.617 $\pm$ 0.039 & \underline{0.609 $\pm$ 0.036} & \underline{0.623 $\pm$ 0.041} \\
   ECG-JEPA & 0.473 $\pm$ 0.019 & 0.456 $\pm$ 0.030 & 0.543 $\pm$ 0.006 \\
   ECGFounder & 0.546 $\pm$ 0.043 & 0.538 $\pm$ 0.048 & 0.580 $\pm$ 0.041 \\
   \textbf{SimDINOv2 Transformer} & \underline{0.683 $\pm$ 0.012} & \underline{0.650 $\pm$ 0.029} & \underline{0.648 $\pm$ 0.032} \\
      \textbf{xECG} & \textbf{0.786 $\pm$ 0.021} & \textbf{0.700 $\pm$ 0.032} & \textbf{0.856 $\pm$ 0.025} \\
    \midrule
    \multicolumn{4}{c}{\textbf{Finetuning}} \\
    \midrule
   ST-MEM & \underline{0.620 $\pm$ 0.034} & \underline{0.618 $\pm$ 0.036} & \underline{0.702 $\pm$ 0.020} \\
   ECG-JEPA & \underline{0.580 $\pm$ 0.014} & \underline{0.576 $\pm$ 0.015} & 0.592 $\pm$ 0.031 \\
   ECGFounder & 0.544 $\pm$ 0.030 & 0.537 $\pm$ 0.031 & 0.579 $\pm$ 0.026 \\
   \textbf{SimDINOv2 Transformer} & \underline{0.620 $\pm$ 0.033} & \underline{0.617 $\pm$ 0.032} & 0.638 $\pm$ 0.035 \\
    \textbf{xECG} & \textbf{0.848 $\pm$ 0.031} & \textbf{0.767 $\pm$ 0.075} & \textbf{0.932 $\pm$ 0.014} \\
    \bottomrule
    \end{tabular}
}
\vspace{0.5em}
\caption{\textbf{Sleep Apnea.} Individual model performance for the Sleep Apnea task (sleep apnea segmentation Apnea-ECG). Each model is trained $5$ times and we report the mean$\pm$ standard deviation. Pairwise differences between models were assessed using two-sided Welch's t-tests across independent runs, with significance defined as $p<0.05$. Bold results indicate best performance across models and underlined indicates 2nd best.}
\label{tab:sleep_apnea_results}
\end{table*}

\begin{table*}[h!]
\centering
{\footnotesize
    \begin{tabular}{l|ccc}
    \toprule
    \multirow{2}{*}{\textbf{Methods}} & \multicolumn{3}{c}{\textbf{DeepBeat PPG}} \\
    & Accuracy& F1 Score& AUROC \\
    \midrule
    \multicolumn{4}{c}{\textbf{Supervised}} \\
    \midrule
   \textbf{Supervised xLSTM} & 0.771 $\pm$ 0.016 & 0.733 $\pm$ 0.012 & 0.876 $\pm$ 0.006 \\
    \midrule
    \multicolumn{4}{c}{\textbf{Linear Probing}} \\
    \midrule
   ST-MEM & \textbf{0.667 $\pm$ 0.048} & \underline{0.560 $\pm$ 0.018} & \underline{0.653 $\pm$ 0.006} \\
   ECG-JEPA & 0.524 $\pm$ 0.010 & 0.448 $\pm$ 0.007 & 0.477 $\pm$ 0.006 \\
   ECGFounder & \textbf{0.714 $\pm$ 0.064} & \underline{0.491 $\pm$ 0.072} & \underline{0.641 $\pm$ 0.019} \\
   \textbf{SimDINOv2 Transf.} & 0.618 $\pm$ 0.034 & \underline{0.562 $\pm$ 0.010} & \underline{0.656 $\pm$ 0.003} \\
   \textbf{xECG} & \textbf{0.701 $\pm$ 0.061} & \textbf{0.616 $\pm$ 0.034} & \textbf{0.751 $\pm$ 0.013} \\
    \midrule
    \multicolumn{4}{c}{\textbf{Finetuning}} \\
    \midrule
   ST-MEM & 0.616 $\pm$ 0.065 & 0.555 $\pm$ 0.030 & 0.643 $\pm$ 0.049 \\
   ECG-JEPA & 0.698 $\pm$ 0.029 & \underline{0.660 $\pm$ 0.022} & 0.818 $\pm$ 0.021 \\
   ECGFounder & \textbf{0.722 $\pm$ 0.048} & \underline{0.686 $\pm$ 0.031} & \underline{0.846 $\pm$ 0.012} \\
   \textbf{SimDINOv2 Transf.} & 0.660 $\pm$ 0.034 & 0.626 $\pm$ 0.020 & 0.780 $\pm$ 0.014 \\
   \textbf{xECG} & \textbf{0.778 $\pm$ 0.024} & \textbf{0.738 $\pm$ 0.018} & \textbf{0.887 $\pm$ 0.017} \\
    \bottomrule
    \end{tabular}
}
\vspace{0.5em}
\caption{\textbf{PPG AF.} Individual model performance for the PPG AF task (AF classification in the PPG dataset DeepBeat). Each model is trained $5$ times and we report the mean$\pm$ standard deviation.  Pairwise differences between models were assessed using two-sided Welch's t-tests across independent runs, with significance defined as $p<0.05$. Bold results indicate best performance across models and underlined indicates 2nd best.}
\label{tab:ppg_results}
\end{table*}

\begin{table*}[h!]
\centering
{\footnotesize
    \begin{tabular}{l|cccc}
    \toprule
    \multirow{2}{*}{\textbf{Methods}} & \multicolumn{4}{c}{\textbf{Age Prediction}} \\
    & MAE (MIMIC-IV)& MAE (PTB-XL)& MAE (CPSC)& MAE (mean) \\
    \midrule
    \multicolumn{5}{c}{\textbf{Supervised}} \\
    \midrule
   \textbf{Supervised xLSTM} & 9.464 $\pm$ 0.124 & 8.229 $\pm$ 0.081 & 8.754 $\pm$ 0.121 & 8.594 $\pm$ 0.089 \\
    \midrule
    \multicolumn{5}{c}{\textbf{Linear Probing}} \\
    \midrule
   ST-MEM & 11.566 $\pm$ 0.025 & 9.804 $\pm$ 0.004 & 11.151 $\pm$ 0.026 & 11.208 $\pm$ 0.016 \\
   ECG-JEPA & 9.980 $\pm$ 0.018 & 8.389 $\pm$ 0.008 & 8.899 $\pm$ 0.017 & 8.809 $\pm$ 0.004 \\
   ECGFounder & \textbf{9.298 $\pm$ 0.019} & \underline{8.098 $\pm$ 0.017} & \underline{8.595 $\pm$ 0.014} & \textbf{8.402 $\pm$ 0.009} \\
   \textbf{SimDINOv2 Transf.} & 10.284 $\pm$ 0.025 & 9.253 $\pm$ 0.025 & 9.959 $\pm$ 0.025 & 9.735 $\pm$ 0.012 \\
   \textbf{xECG} & \underline{9.334 $\pm$ 0.011} & \textbf{7.989 $\pm$ 0.006} & \textbf{8.489 $\pm$ 0.006} & \underline{8.659 $\pm$ 0.006} \\
    \midrule
    \multicolumn{5}{c}{\textbf{Finetuning}} \\
    \midrule
   ST-MEM & \textbf{8.247 $\pm$ 0.087} & \textbf{7.393 $\pm$ 0.123} & \textbf{7.245 $\pm$ 0.133} & \textbf{7.225 $\pm$ 0.084} \\
   ECG-JEPA & \underline{8.849 $\pm$ 0.123} & 7.927 $\pm$ 0.195 & 8.169 $\pm$ 0.118 & \underline{7.829 $\pm$ 0.100} \\
   ECGFounder & 9.061 $\pm$ 0.139 & \underline{7.884 $\pm$ 0.223} & 8.260 $\pm$ 0.168 & 7.970 $\pm$ 0.128 \\
   \textbf{SimDINOv2 Transf.} & 9.866 $\pm$ 0.069 & 8.664 $\pm$ 0.068 & 9.303 $\pm$ 0.073 & 8.932 $\pm$ 0.032 \\
   \textbf{xECG} & \underline{8.818 $\pm$ 0.040} & \underline{7.670 $\pm$ 0.036} & \underline{7.943 $\pm$ 0.068} & \underline{7.790 $\pm$ 0.028} \\
    \bottomrule
    \end{tabular}
}
\vspace{0.5em}
\caption{\textbf{Age.} Individual model mean absolute error (MAE) for the Age task (age regression, trained on CODE-15\% and evaluated on MIMIC-IV-ECG, PTB-XL and CPSC2018). Each model is trained $5$ times and we report the mean$\pm$ standard deviation.  Pairwise differences between models were assessed using two-sided Welch's t-tests across independent runs, with significance defined as $p<0.05$. Bold results indicate best performance across models and underlined indicates 2nd best.}

\label{tab:age_results}
\end{table*}
\begin{table*}[h!]
\centering
{\footnotesize
    \begin{tabular}{l|cccc}
    \toprule
    \multirow{2}{*}{\textbf{Methods}} & \multicolumn{4}{c}{\textbf{Age Prediction}} \\
    & SMAPE (MIMIC-IV)& SMAPE (PTB-XL)& SMAPE (CPSC)& SMAPE (mean) \\
    \midrule
    \multicolumn{5}{c}{\textbf{Supervised}} \\
    \midrule
   \textbf{Sup. xLSTM} & 0.079 $\pm$ 0.001 & 0.075 $\pm$ 0.001 & 0.082 $\pm$ 0.001 & 0.079 $\pm$ 0.001 \\
    \midrule
    \multicolumn{5}{c}{\textbf{Linear Probing}} \\
    \midrule
   ST-MEM & 0.098 $\pm$ 0.000 & 0.089 $\pm$ 0.000 & 0.104 $\pm$ 0.000 & 0.105 $\pm$ 0.000 \\
   ECG-JEPA & 0.083 $\pm$ 0.000 & 0.076 $\pm$ 0.000 & 0.083 $\pm$ 0.000 & 0.081 $\pm$ 0.000 \\
   ECGFounder & \textbf{0.077 $\pm$ 0.000} & \underline{0.073 $\pm$ 0.000} & \underline{0.081 $\pm$ 0.000} & \textbf{0.078 $\pm$ 0.000} \\
   \textbf{SimDINOv2 Trans.} & 0.086 $\pm$ 0.000 & 0.085 $\pm$ 0.000 & 0.094 $\pm$ 0.000 & 0.091 $\pm$ 0.000 \\
   \textbf{xECG} & \underline{0.078 $\pm$ 0.000} & \textbf{0.073 $\pm$ 0.000} & \textbf{0.080 $\pm$ 0.000} & \underline{0.081 $\pm$ 0.000} \\
    \midrule
    \multicolumn{5}{c}{\textbf{Finetuning}} \\
    \midrule
   ST-MEM & \textbf{0.069 $\pm$ 0.001} & \textbf{0.067 $\pm$ 0.001} & \textbf{0.068 $\pm$ 0.001} & \textbf{0.067 $\pm$ 0.001} \\
   ECG-JEPA & \underline{0.073 $\pm$ 0.001} & \underline{0.070 $\pm$ 0.002} & \underline{0.075 $\pm$ 0.001} & \underline{0.071 $\pm$ 0.001} \\
   ECGFounder & 0.075 $\pm$ 0.001 & \underline{0.072 $\pm$ 0.002} & 0.078 $\pm$ 0.001 & 0.074 $\pm$ 0.001 \\
   \textbf{SimDINOv2 Trans.} & 0.083 $\pm$ 0.001 & 0.079 $\pm$ 0.001 & 0.088 $\pm$ 0.001 & 0.083 $\pm$ 0.000 \\
   \textbf{xECG} & \underline{0.073 $\pm$ 0.000} & \underline{0.070 $\pm$ 0.000} & \underline{0.075 $\pm$ 0.001} & \underline{0.072 $\pm$ 0.000} \\
    \bottomrule
    \end{tabular}
}
\vspace{0.5em}
\caption{\textbf{Age.} Individual model symmetric mean absolute percentage error (SMAPE) for the Age task (age regression, trained on CODE-15\% and evaluated on MIMIC-IV-ECG, PTB-XL and CPSC2018). Each model is trained $5$ times and we report the mean$\pm$ standard deviation.  Pairwise differences between models were assessed using two-sided Welch's t-tests across independent runs, with significance defined as $p<0.05$. Bold results indicate best performance across models and underlined indicates 2nd best.}
\label{tab:age_results}
\end{table*}

\begin{table*}[h!]
\centering
{\footnotesize
    \begin{tabular}{l|ccc}
    \toprule
    \multirow{2}{*}{\textbf{Methods}} & \multicolumn{3}{c}{\textbf{Blood test}} \\
    & Accuracy & F1 Score & AUROC \\
    \midrule
    \multicolumn{4}{c}{\textbf{Supervised}} \\
    \midrule
   \textbf{Supervised xLSTM} & 0.425 $\pm$ 0.004 & 0.421 $\pm$ 0.003 & 0.683 $\pm$ 0.001 \\
    \midrule
    \multicolumn{4}{c}{\textbf{Linear Probing}} \\
    \midrule
   ST-MEM & 0.424 $\pm$ 0.000 & 0.416 $\pm$ 0.000 & 0.708 $\pm$ 0.000 \\
   ECG-JEPA & 0.432 $\pm$ 0.000 & 0.424 $\pm$ 0.000 & 0.722 $\pm$ 0.000 \\
   ECGFounder & 0.430 $\pm$ 0.000 & 0.424 $\pm$ 0.000 & 0.711 $\pm$ 0.000 \\
   \textbf{SimDINOv2 Transformer} & \underline{0.437 $\pm$ 0.000} & \underline{0.431 $\pm$ 0.000} & \underline{0.724 $\pm$ 0.001} \\
   \textbf{xECG} & \textbf{0.439 $\pm$ 0.000} & \textbf{0.435 $\pm$ 0.001} & \textbf{0.733 $\pm$ 0.000} \\
    \midrule
    \multicolumn{4}{c}{\textbf{Finetuning}} \\
    \midrule
   ST-MEM & \textbf{0.456 $\pm$ 0.002} & \textbf{0.457 $\pm$ 0.002} & \underline{0.744 $\pm$ 0.002} \\
   ECG-JEPA & 0.440 $\pm$ 0.000 & 0.434 $\pm$ 0.001 & 0.735 $\pm$ 0.001 \\
   ECGFounder & 0.440 $\pm$ 0.002 & 0.437 $\pm$ 0.003 & 0.714 $\pm$ 0.003 \\
   \textbf{SimDINOv2 Transformer} & 0.445 $\pm$ 0.004 & 0.443 $\pm$ 0.005 & 0.725 $\pm$ 0.001 \\
   \textbf{xECG} & \underline{0.451 $\pm$ 0.001} & \underline{0.450 $\pm$ 0.002} & \textbf{0.747 $\pm$ 0.002} \\
    \bottomrule
    \end{tabular}
}
\vspace{0.5em}
\caption{\textbf{Blood test.} Individual model performance for the Blood test task ((ab)normal classification of blood test results from MIMIC-IV-ECG). Each model is trained $5$ times and we report the mean$\pm$ standard deviation. Pairwise differences between models were assessed using two-sided Welch's t-tests across independent runs, with significance defined as $p<0.05$. Bold results indicate best performance across models and underlined indicates 2nd best.}
\label{tab:lab_test_results}
\end{table*}

\begin{table*}[h!]
\centering
{\footnotesize
    \begin{tabular}{l|ccc}
    \toprule
    \multirow{2}{*}{\textbf{Methods}} & \multicolumn{3}{c}{\textbf{Survival Analysis}} \\
    & CI (CODE-15, Val)& CI (MIMIC-IV-ECG, Test)& CI (difference) \\
    \midrule
    \multicolumn{4}{c}{\textbf{Supervised}} \\
    \midrule
   \textbf{Supervised xLSTM} & 0.650 $\pm$ 0.018 & 0.630 $\pm$ 0.016 & 0.020 $\pm$ 0.015 \\
    \midrule
    \multicolumn{4}{c}{\textbf{Linear Probing}} \\
    \midrule
   ST-MEM & 0.755 $\pm$ 0.005 & 0.697 $\pm$ 0.001 & \textbf{0.058 $\pm$ 0.006} \\
   ECG-JEPA & \textbf{0.830 $\pm$ 0.004} & \textbf{0.706 $\pm$ 0.001} & 0.124 $\pm$ 0.004 \\
   ECGFounder & \underline{0.817 $\pm$ 0.004} & 0.683 $\pm$ 0.001 & 0.134 $\pm$ 0.004 \\
   \textbf{SimDINOv2 Transformer} & 0.779 $\pm$ 0.005 & 0.695 $\pm$ 0.001 & \underline{0.084 $\pm$ 0.005} \\
   \textbf{xECG} & \underline{0.811 $\pm$ 0.005} & \underline{0.703 $\pm$ 0.001} & 0.108 $\pm$ 0.005 \\
    \midrule
    \multicolumn{4}{c}{\textbf{Finetuning}} \\
    \midrule
   ST-MEM & 0.774 $\pm$ 0.003 & 0.694 $\pm$ 0.004 & \textbf{0.080 $\pm$ 0.000} \\
   ECG-JEPA & \textbf{0.827 $\pm$ 0.004} & \underline{0.706 $\pm$ 0.001} & 0.122 $\pm$ 0.005 \\
   ECGFounder & \textbf{0.822 $\pm$ 0.005} & 0.688 $\pm$ 0.002 & 0.134 $\pm$ 0.003 \\
   \textbf{SimDINOv2 Transformer} & 0.785 $\pm$ 0.007 & \underline{0.701 $\pm$ 0.004} & \textbf{0.084 $\pm$ 0.008} \\
   \textbf{xECG} & 0.817 $\pm$ 0.005 & \textbf{0.710 $\pm$ 0.001} & 0.107 $\pm$ 0.005 \\
    \bottomrule
    \end{tabular}
}
\vspace{0.5em}
\caption{\textbf{Mortality.} Concordance Index (CI) for individual models for the Mortality task (survival analysis for mortality prediction, trained on CODE-15\% and evaluated on MIMIC-IV-ECG). Each model is trained $5$ times and we report the mean$\pm$ standard deviation. Pairwise differences between models were assessed using two-sided Welch's t-tests across independent runs, with significance defined as $p<0.05$. Bold results indicate best performance across models and underlined indicates 2nd best.}
\label{tab:mortality_results}
\end{table*}

\newpage
\clearpage

\subsection*{Downstream tasks hyperparameters}

This section details the hyperparameters and data processing procedures used for linear probing and fine-tuning on all downstream tasks. It is important to note that ECGFounder, as a convolutional-based model, does not utilize drop path or layer-wise learning rate decay. No data augmentations were applied to any downstream task, with the exception of the DeepBeat PPG dataset, which was only available in a pre-augmented format for the training split.
For all downstream tasks, we used the AdamW optimiser with a learning rate scheduler. The first epoch was used for learning rate warm-up, after which the learning rate was progressively reduced to zero following a cosine schedule by the final epoch.

Patch representation is specific to our transformer and xLSTM-based models (xECG, SimDINOv2, and Supervised xLSTM). This hyperparameter determines the pooling method applied to patch embeddings to generate the final signal representation. We employed two pooling strategies: average pooling (avg), which computes the mean of all patch embeddings, and max pooling (max), which takes the maximum value across the sequence for each feature dimension.
For all the tasks and all the methods we selected the best model on the validation set on the metric used for the BenchECG score.

\textbf{PTB-XL.} For the PTB-XL dataset, a batch size of 256 was used for linear probing across all models. During fine-tuning, the batch size was reduced for certain models due to memory constraints. Table \ref{tab:hyperparameters_ptbxl} provides a comprehensive list of all hyperparameters for this task.

\begin{table*}[h!]
\centering
{\footnotesize
    \begin{tabular}{l|cccccc}
    \toprule
    \multirow{3}{*}{\textbf{Hyperparameter}} & \multicolumn{6}{c}{\textbf{PTB-XL}} \\
     & ECGFounder& ECG-JEPA& ST-MEM & \makecell{SimDINOv2\\Transf.}& xECG& \makecell{Sup.\\xLSTM}\\
    \midrule
    \multicolumn{7}{c}{Linear Probing} \\
    \midrule
    learning rate head & 0.001 & 0.001 & 0.001 & 0.001 & 0.001 & - \\
    patch representation & - & - & - & avg & avg  & - \\
    \midrule
    \multicolumn{7}{c}{Finetuning} \\
    \midrule
    learning rate head & 0.001& 0.001& 0.0003& 0.001& 0.001& 0.0001\\
    learning rate encorder & 0.00001 & 0.00001 & 0.0003 & 0.00001 & 0.00003 & 0.0001\\
    layerwise lr decay&- & 0.75 & 0.75 & 0.75 & 0.75 & -\\
    drop path & -& 0.5 & 0.5 & -& 0.5 & -\\
    batch size& 256 & 96 & 128 & 256 & 256 & 256 \\
    patch representation & - & - & - & avg & avg  & avg \\
    \bottomrule
    \end{tabular}
}
\vspace{0.5em}
\caption{Hyperparameters for the PTB-XL task.}
\label{tab:hyperparameters_ptbxl}
\end{table*}

\textbf{CPSC2018.} In the nine-class multi-label classification task on the CPSC2018 dataset, all models were trained for 30 epochs with a weight decay of 0.1. The learning rates varied depending on the specific model. A batch size of 256 was used for linear probing, while for fine-tuning, it was adjusted for some models to accommodate memory limitations. Because this dataset contain signals with different lenghts, spanning from 7 to 60 seconds we cropped signals to 10 seconds and applied padding if they were smaller than that.
The complete hyperparameter settings for this task are presented in Table \ref{tab:hyperparameters_cpsc}.

\begin{table*}[h!]
\centering
{\footnotesize
    \begin{tabular}{l|cccccc}
    \toprule
    \multirow{3}{*}{\textbf{Hyperparameter}} & \multicolumn{6}{c}{\textbf{CPSC2018}} \\
     & ECGFounder& ECG-JEPA& ST-MEM & \makecell{SimDINOv2\\Transf.}& xECG& \makecell{Sup.\\xLSTM}\\
    \midrule
    \multicolumn{7}{c}{Linear Probing} \\
    \midrule
    learning rate head & 0.001 & 0.001 & 0.001 & 0.001 & 0.001 & - \\
    final normalisation  & - & - & - & batch & -  & - \\
    patch representation & - & - & - & max& max& - \\
    \midrule
    \multicolumn{7}{c}{Finetuning} \\
    \midrule
    learning rate head & 0.001& 0.001& 0.0003& 0.001& 0.001& 0.0001\\
    learning rate encorder& 0.00001& 0.00001& 0.0003& 0.00001& 0.00003& 0.0001\\
    layerwise lr decay& -& 0.75& 0.75& 0.75& 0.75& -\\
    drop path & -& 0.5& 0.5& -& 0.5& -\\
    batch size& 256& 96& 128& 256& 128& 128\\
    final normalisation  & - & - & - & batch norm& -  & - \\
    patch representation & - & - & - & max& max& max\\
    \bottomrule
    \end{tabular}
}
\vspace{0.5em}
\caption{Hyperparameters for classification task on CPSC2018.}
\label{tab:hyperparameters_cpsc}
\end{table*}

\textbf{MIT-BIH.} For the beat-level classification task on the MIT-BIH dataset, the training procedures differed significantly across the various methods, reflecting the R-peak level annotations. For the transformer and xLSTM-based models, which have sufficiently small patch sizes (ECG-JEPA: 20ms, ST-MEM: 300ms, our baselines: 250ms), we adopted a patch-level classification approach. This ensures that each patch contains at most one heartbeat. During training, we utilised overlapping windows centered around the R-peak, with the window length being model-dependent. For validation and testing, non-overlapping windows were used.  Our recurrent baselines processed the entire signal as a single sample.
ECGFounder, due to its architecture, does not support a fine-grained, beat-level output. Therefore, this model was trained by feeding it single heartbeats, each within a 1-second window centered on the corresponding R-peak.
Because this dataset consists of signals recorded with a two leads configuration, the missing leads are set to zero for the whole duration of the ecg.
For both linear probing and fine-tuning on this dataset, models were trained for 20 epochs. A consistent learning rate of 0.001 was used for linear probing, while different learning rates were employed for fine-tuning. The batch size for linear probing was 256 and was adjusted during fine-tuning based on memory availability. All hyperparameters for the MIT-BIH classification task are detailed in Table \ref{tab:hyperparameters_mit_cls} .

\textbf{R-peak.} For the R-peak detection task, a similar windowing approach was utilised. However, a key distinction was the use of non-overlapping windows during the training phase for all models. For our recurrent models, this non-overlapping window strategy was also applied during evaluation, instead of processing the entire 30-minute signal.
ST-MEM and ECG-JEPA output 12 and 8 patch embeddings, respectively, for the same segment of ECG originally divided into patches. We then use the patch embedding from lead II as the input to the detection head.
To enable detection at a signal-pixel level, a detection head was appended to the models. For the transformer and xLSTM-based architectures, this head produces an output with a dimensionality matching the input patch size, thereby allowing for fine-grained temporal localization. In the case of ECGFounder, which lacks inherent spatial correspondence in its output due to its convolutional nature, a prediction head was applied that matched the size of its accepted input window. For both linear probing and fine-tuning on this dataset, models were trained for 30 epochs. All the hyperparameters are shown in Table \ref{tab:hyperparameters_mit_r_peak}

\begin{table*}[h!]
\centering
{\footnotesize
    \begin{tabular}{l|cccccc}
    \toprule
    \multirow{3}{*}{\textbf{Hyperparameter}} & \multicolumn{6}{c}{\textbf{MIT-BIH (classification)}} \\
     & ECGFounder& ECG-JEPA& ST-MEM & \makecell{SimDINOv2\\Transf.}& xECG& \makecell{Sup.\\xLSTM}\\
    \midrule
    \multicolumn{7}{c}{Linear Probing} \\
    \midrule
    learning rate head & 0.001 & 0.001 & 0.001 & 0.001 & 0.001 & - \\
    weight decay & 0.1 & 0.1 & 0.1 & 0.1 & 0.1  & - \\
    window len & 10s&	10s&	10s&	10s& 	36s& 	- \\
 patch representation & - & - & - & avg & avg  &- \\
    \midrule
    \multicolumn{7}{c}{Finetuning} \\
    \midrule
    learning rate head & 0.001& 0.001& 0.001& 0.01& 0.001& 0.0001\\
    learning rate encorder& 0.00001& 0.00001& 0.001& 0.00001& 0.000001& 0.0001\\
    weight decay& 0.1& 0.1& 0.1& 0.1& 0.0& 0.0\\
    layerwise lr decay& -& 0.75& 0.75& 0.75& 0.75& -\\
    drop path & 0.0& 0.5& 0.5& 0.5& 0.0& 0.0\\
    batch size& 256& 96& 96& 96& 64& 64\\
    window len & 10s&	10s&	10s&	10s& 	36s& 	36s\\
    \end{tabular}
}
\vspace{0.5em}
\caption{Hyperparameters for each group.}
\label{tab:hyperparameters_mit_cls}
\end{table*}

\begin{table*}[h!]
\centering
{\footnotesize
    \begin{tabular}{l|cccccc}
    \toprule
    \multirow{3}{*}{\textbf{Hyperparameter}} & \multicolumn{6}{c}{\textbf{MIT-BIH (R-peak detection)}} \\
     & ECGFounder& ECG-JEPA& ST-MEM & \makecell{SimDINOv2\\Transf.}& xECG& \makecell{Sup.\\xLSTM}\\
    \midrule
    \multicolumn{7}{c}{Linear Probing} \\
    \midrule
    learning rate head & 0.001& 0.01& 0.001& 0.01& 0.01& - \\
    weight decay & 0.1& 0.1& 0.1& 0.1& 0.1& - \\
    window len & 10s&	10s&	10s&	10s& 	1m12s& 	- \\
    patch normalisation  & - & - & - & layer norm & -  & - \\
    \midrule
    \multicolumn{7}{c}{Finetuning} \\
    \midrule
    learning rate head & 0.001& 0.001& 0.01& 0.01& 0.001& 0.01\\
    learning rate encorder& 0.00001& 0.001& 0.01& 0.001& 0.001& 0.01\\
    weight decay& 0.1& 0.1& 0.1& 0.1& 0.0& 0.0\\
    layerwise lr decay& -& 0.75& 0.75& 0.75& 0.75& 0.75\\
    drop path & -& 0.5& 0.5& 0.5& -& 0.0\\
    batch size& 256& 96& 96& 96& 64& 64\\
    window len & 10s&	10s&	10s&	10s& 	1m12s& 	1m12s\\
    \bottomrule
    \end{tabular}
}
\vspace{0.5em}
\caption{Hyperparameters for each group.}
\label{tab:hyperparameters_mit_r_peak}
\end{table*}

\textbf{Exercise.} On the ECG in High Intensity Dataset we followed the same procedure of the R-peak detection task of MIT-BIH. The only difference is that in this dataset each recording is of 20 seconds so window sizes matches the 20 seconds for our recurrent models and 10 seconds for all the others. Training was done for 30 epochs and a batch size of 8 (due to the very small amount of data). All the other parameters are shown in Table \ref{tab:hyperparameters_exe_peak}.
The ECGs in this dataset consist only of lead II recordings, thus we padded to zero the rest of the leads for all the model except for ECGFounder where another version of the model was released accepting one single lead.

\begin{table*}[h!]
\centering
{\footnotesize
    \begin{tabular}{l|cccccc}
    \toprule
    \multirow{3}{*}{\textbf{Hyperparameter}} & \multicolumn{6}{c}{\textbf{ECG in Intense Exercise Dataset (R-peak detection)}} \\
     & ECGFounder& ECG-JEPA& ST-MEM & \makecell{SimDINOv2\\Transf.}& xECG& \makecell{Sup.\\xLSTM}\\
    \midrule
    \multicolumn{7}{c}{Linear Probing} \\
    \midrule
    learning rate head & 0.001& 0.1& 0.01& 0.1& 0.1& - \\
    weight decay & 0.1& 0.1& 0.1& 0.1& 0.1& - \\
    window len & 10s&	10s&	10s&	10s& 	20s& 	- \\
    patch normalisation  & - & - & - & layer norm & -  & - \\
    \midrule
    \multicolumn{7}{c}{Finetuning} \\
    \midrule
    learning rate head & 0.001& 0.01& 0.01& 0.1& 0.01& 0.001\\
    learning rate encorder& 0.00001& 0.001& 0.01& 0.001& 0.001& 0.001\\
    weight decay& 0.1& 0.1& 0.1& 0.1& 0.0& 0.0\\
    layerwise lr decay& -& 0.75& 0.75& 0.75& 0.75& -\\
    drop path & -& 0.5& 0.5& 0.5& 0.5& 0.0\\
    window len & 10s&	10s&	10s&	10s& 	20s& 	20s\\
    \bottomrule
    \end{tabular}
}
\vspace{0.5em}
\caption{Hyperparameters for each group.}
\label{tab:hyperparameters_exe_peak}
\end{table*}

\textbf{Sleep Apnea.} Training on the Sleep Apnea-ECG dataset required different approaches for recurrent and non-recurrent models due to the minute-level annotations. For non-recurrent models that accept 10-second inputs, each overnight signal was divided into non-overlapping 10-second segments, each inheriting the label of its corresponding minute. A linear head on the final signal representation was used for prediction, and the loss was calculated for each 10-second segment during training. For evaluation, predictions from the six segments within each minute were aggregated and averaged. Conversely, our recurrent models (xECG and Supervised xLSTM) were fed 3-minute signal segments to leverage their ability to process longer sequences. A prediction head was appended to each patch, and the loss was applied at the patch level. During validation and testing, these patch-level predictions were averaged at the minute level to align with the ground-truth labels. All models were trained for 20 epochs with batch sizes adjusted based on memory consumption. The corresponding hyperparameters are shown in Table \ref{tab:hyperparameters_sleep}.
The ECGs in this dataset consist only of lead II recordings, thus we padded to zero the rest of the leads for all the model except for ECGFounder where the single-lead configuration was utilised.

\begin{table*}[h!]
\centering
{\footnotesize
    \begin{tabular}{l|cccccc}
    \toprule
    \multirow{3}{*}{\textbf{Hyperparameter}} & \multicolumn{6}{c}{\textbf{Sleep Apnea-ECG}} \\
     & ECGFounder& ECG-JEPA& ST-MEM & \makecell{SimDINOv2\\Transf.}& xECG& \makecell{Sup.\\xLSTM}\\
    \midrule
    \multicolumn{7}{c}{Linear Probing} \\
    \midrule
    learning rate head & 0.001& 0.001& 0.001& 0.0001& 0.001& - \\
    weight decay & 0.1& 0.1& 0.1& 0.1& 0.1& - \\
    batch size& 256& 256& 256& 256& 3&\\
    window len & 10s&	10s&	10s&	10s& 	3m& 	- \\
    patch normalisation  & - & - & - & layer norm & -  & - \\
    patch representation & - & - & - & avg & max  &- \\
    \midrule
    \multicolumn{7}{c}{Finetuning} \\
    \midrule
    learning rate head & 0.001& 0.001& 0.0003& 0.00001& 0.001& 0.0001\\
    learning rate encorder& 0.000001& 0.000001& 0.0003& 0.000001& 0.0001& 0.0001\\
    weight decay& 0.1& 0.1& 0.1& 0.1& 0.1& 0.0\\
    layerwise lr decay& -& 0.75& 0.75& 0.75& 0.9& -\\
    drop path & -& 0.5& 0.5& 0.5& 0.2& -\\
    batch size& 256& 96& 128& 128& 3&3\\
    window len & 10s&	10s&	10s&	10s& 	3m& 	3m\\

     patch representation & - & - & - & avg & max  & avg\\
    \bottomrule
    \end{tabular}
}
\vspace{0.5em}
\caption{Hyperparameters for each group.}
\label{tab:hyperparameters_sleep}
\end{table*}

\textbf{PPG AF.} For the Atrial Fibrillation classification on the DeepBeat PPG dataset, all models were trained for 30 epochs using the maximum batch size that each method could use to accelerate training. As we did not have access to the original, non-augmented data, we adopted a strategy of training on a randomly selected 10\% subset of the training set for each epoch. All hyperparameters for this task are detailed in Table \ref{tab:hyperparameters_ppg}.
The recordings in this dataset are single-channel, which we treated as lead II recordings. For all models except ECGFounder, we zero-padded the remaining leads, while for ECGFounder we used the single-lead configuration.

\begin{table*}[h!]
\centering
{\footnotesize
    \begin{tabular}{l|cccccc}
    \toprule
    \multirow{3}{*}{\textbf{Hyperparameter}} & \multicolumn{6}{c}{\textbf{DeepBeat PPG}} \\
     & ECGFounder& ECG-JEPA& ST-MEM & \makecell{SimDINOv2\\Transf.}& xECG& \makecell{Sup.\\xLSTM}\\
    \midrule
    \multicolumn{7}{c}{Linear Probing} \\
    \midrule
    learning rate head & 0.1& 0.001& 0.0003& 0.001& 0.001& - \\
    weight decay & 0.1& 0.1& 0.1& 0.1& 0.1& - \\
 batch size& 512& 96& 128& 256& 512&\\
    patch normalisation  & - & - & - & layer norm & -  & - \\
 patch representation & - & - & - & avg & avg  &- \\
    \midrule
    \multicolumn{7}{c}{Finetuning} \\
    \midrule
    learning rate head & 0.01& 0.001& 0.0003& 0.001& 0.001& 0.0001\\
    learning rate encorder& 0.001& 0.00001& 0.0003& 0.0001& 0.00001& 0.0001\\
    weight decay& 0.1& 0.1& 0.1& 0.1& 0.1& 0.1\\
    layerwise lr decay& -& 0.75& 0.75& 0.75& 1.0& -\\
    drop path & -& 0.5& 0.5& 0.5& -& -\\
 batch size& 128& 96& 128& 256& 128&128\\
 patch normalisation  & - & - & - & layer norm & -  &- \\
 patch representation & - & - & - & avg & avg  &avg\\
    \end{tabular}
}
\vspace{0.5em}
\caption{Hyperparameters for each group.}
\label{tab:hyperparameters_ppg}
\end{table*}

\textbf{Age.} On the age regression task, all models were trained for 15 epochs. The batch size for linear probing was 512, while for fine-tuning, it was reduced based on memory limits. The hyperparameters are presented in Table \ref{tab:hyperparameters_age}.

\begin{table*}[h!]
\centering
{\footnotesize
    \begin{tabular}{l|cccccc}
    \toprule
    \multirow{3}{*}{\textbf{Hyperparameter}} & \multicolumn{6}{c}{\textbf{Age regression}} \\
     & ECGFounder& ECG-JEPA& ST-MEM & \makecell{SimDINOv2\\Transf.}& xECG& \makecell{Sup.\\xLSTM}\\
    \midrule
    \multicolumn{7}{c}{Linear Probing} \\
    \midrule
    learning rate head & 0.1& 0.1& 0.01& 0.01& 0.01& - \\
    weight decay & 0.01& 0.01& 0.01& 0.01& 0.01& - \\
 patch representation & - & - & - & avg & avg  &- \\
    \midrule
    \multicolumn{7}{c}{Finetuning} \\
    \midrule
    learning rate head & 0.1& 0.1& 0.0003& 0.01& 0.01& 0.001\\
    learning rate encorder& 0.01& 0.01& 0.01& 0.01& 0.1& 0.01\\
    weight decay& 0.01& 0.01& 0.01& 0.01& 0.1& 0.01\\
    layerwise lr decay& -& 0.75& 0.75& 0.75& 0.75& 0.75\\
    drop path & -& 0.5& -& -& 0.2& 0.5\\
 batch size& 512& 96& 128& 512& 512&512\\
 patch representation & - & - & - & avg & avg  &avg\\
    \end{tabular}
}
\vspace{0.5em}
\caption{Hyperparameters for each group.}
\label{tab:hyperparameters_age}
\end{table*}

\textbf{Blood test.} For the blood test abnormality detection task, all the models were trained for 20 epochs. Batch size for linear probing was set to 512, while for fine-tuning, it was reduced based on memory limits. Hyperparameters are presented in Table \ref{tab:hyperparameters_blood}

\begin{table*}[h!]
\centering
{\footnotesize
    \begin{tabular}{l|cccccc}
    \toprule
    \multirow{3}{*}{\textbf{Hyperparameter}} & \multicolumn{6}{c}{\textbf{Blood test}} \\
     & ECGFounder& ECG-JEPA& ST-MEM & \makecell{SimDINOv2\\Transf.}& xECG& \makecell{Sup.\\xLSTM}\\
    \midrule
    \multicolumn{7}{c}{Linear Probing} \\
    \midrule
    learning rate head & 0.001& 0.003& 0.01& 0.001& 0.001& - \\
    weight decay & 0.1& 0.1& 0.1& 0.1& 0.1& - \\
 final normalisation  & -& -& -& batch norm& -&-\\
 patch representation & - & - & - & avg & avg  &- \\
    \midrule
    \multicolumn{7}{c}{Finetuning} \\
    \midrule
    learning rate head & 0.001& 0.001& 0.001& 0.001& 0.001& 0.0001\\
    learning rate encorder& 0.0001& 0.00001& 0.001& 0.00001& 0.00003& 0.0001\\
    weight decay& 0.1& 0.1& 0.1& 0.1& 0.1& 0.1\\
    layerwise lr decay& -& 0.75& 0.75& 0.75& 0.75& -\\
    drop path & -& 0.5& -& -& 0.2& 0.5\\
 batch size& 512& 96& 96& 512& 512&512\\
 patch representation & - & - & - & avg & avg  &avg\\
    \end{tabular}
}
\vspace{0.5em}
\caption{Hyperparameters for each group.}
\label{tab:hyperparameters_blood}
\end{table*}

\textbf{Mortality.} For the mortality risk task, all the models were trained for 30 epochs. Batch size for linear probing was set to 512, while for fine-tuning, it was reduced based on memory limits. Hyperparameters are presented in Table \ref{tab:hyperparameters_mortality}

\begin{table*}[h!]
\centering
{\footnotesize
    \begin{tabular}{l|cccccc}
    \toprule
    \multirow{3}{*}{\textbf{Hyperparameter}} & \multicolumn{6}{c}{\textbf{Mortality}} \\
     & ECGFounder& ECG-JEPA& ST-MEM & \makecell{SimDINOv2\\Transf.}& xECG& \makecell{Sup.\\xLSTM}\\
    \midrule
    \multicolumn{7}{c}{Linear Probing} \\
    \midrule
    learning rate head & 0.001& 0.001& 0.01& 0.001& 0.001& - \\
    weight decay & 0.1& 0.1& 0.1& 0.1& 0.1& - \\
 final normalisation  & -& -& -& batch norm& -&-\\
 patch representation & - & - & - & avg & avg  &- \\
    \midrule
    \multicolumn{7}{c}{Finetuning} \\
    \midrule
    learning rate head & 0.001& 0.001& 0.0003& 0.001& 0.001& 0.001\\
    learning rate encorder& 0.0001& 0.000001& 0.00003& 0.00001& 0.000001& 0.0001\\
    weight decay& 0.1& 0.1& 0.1& 0.1& 0.1& 0.1\\
    layerwise lr decay& -& 0.75& 0.75& 0.75& 0.75& -\\
    drop path & -& 0.5& 0.0& 0.0& 0.0& -\\
 batch size& 512& 96& 96& 512& 512&512\\
 final normalisation  & -& -& -& batch norm& -&-\\
 patch representation & - & - & - & avg & avg  &avg\\
    \end{tabular}
}
\vspace{0.5em}
\caption{Hyperparameters for each group.}
\label{tab:hyperparameters_mortality}
\end{table*}

\subsection*{Pre-training hyperparameters}

Our xECG model was pretrained using the SimDINOv2 framework for 100 epochs with a batch size of 512. The learning rate was initialised to $0.0001$ and managed by a cosine annealing scheduler, which included a five-epoch linear warm-up phase. We applied a layer-wise learning rate decay of $0.9$, and gradient clipping was set to a maximum norm of $3.0$. Both weight decay and the teacher model's exponential moving average (EMA) momentum ($\lambda_t$) were scheduled. Weight decay increased linearly from 0.04 to 0.4 over the course of training, while the EMA momentum was scheduled to increase from 0.99 to 1.0 by the final epoch. 

For the self-distillation process, we generated two global views (random crops of $80\%$ of the signal length) and four local views (random crops of $40\%$) for each input sample. Patches within the student model's input views were masked with a probability of $0.3$.
Random lead dropout was used with a probability of $0.2$ for all leads except for lead II. Multiplicative Gaussian jitter (amplitude $A$=$0.6$) and random amplitude scaling (range $R=0.2$) were each applied with a probability of $0.1$.

The xECG model operates on signals sampled at a frequency ($f_m$) of $100$ Hz. The input signal is divided into patches, where each patch consists of $25$ time points, corresponding to a duration of $250$ ms. The transformer baseline was pretrained using the same set of hyperparameters described above. The sole architectural difference was the inclusion of learnable positional embeddings, which are necessary for the transformer but not for our recurrent xECG model.

\end{document}